\documentclass{article}
\usepackage[utf8]{inputenc}

\usepackage{apacite}
\usepackage{amsmath}
\usepackage{amsthm}
\usepackage{booktabs}
\usepackage{multirow}
\usepackage{titlepic}
\usepackage{graphicx}
\usepackage{color}
\usepackage{tikz}
\usepackage{amsfonts}

\tikzstyle{seen}=[fill=lightgray, draw=black, shape=circle, minimum size=0.4cm, inner sep=0.1cm]
\tikzstyle{unseen}=[fill=white, draw=black, shape=circle, minimum size=0.4cm, inner sep=0.1cm]
\tikzstyle{none}=[]
\tikzstyle{full}=[->, draw=black]
\tikzstyle{dashed}=[->, draw=black]

\theoremstyle{definition}
\newtheorem{Def}{Definition}
\newtheorem{definition}[Def]{Definition}

\newtheorem{lemma}[Def]{Lemma}

\newtheorem{example}[Def]{Example}

\newcommand{\Gardenfors}{G\"ardenfors}

\newcommand{\z}{\mathbf{z}}
\newcommand{\Z}{\mathbf{Z}}
\newcommand{\x}{\mathbf{x}}
\newcommand{\X}{\mathbf{X}}
\newcommand{\con}{\mathbf{c}}

\newcommand{\cc}{c} 
\newcommand{\crc}{c} 
\newcommand{\czero}{\mathbf{c_0}}
\newcommand{\cone}{\mathbf{c_1}}
\newcommand{\ctwo}{\mathbf{c_2}}
\newcommand{\cthree}{\mathbf{c_3}}

\title{The Conceptual VAE}
\author{Razin A. Shaikh, Sara Sabrina Zemlji\v{c}, Sean Tull and Stephen Clark\\
Cambridge Quantum / Quantinuum\\ 17 Beaumont Street, Oxford, UK\\
\small{\texttt{\{razin.shaikh,sara.zemljic,sean.tull,steve.clark\}@cambridgequantum.com}}}

\titlepic{\includegraphics[width=1cm]{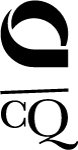}}
\date{21 March 2022}

\usepackage{amsmath}
\usepackage{amsthm}
\usepackage{amssymb}
\usepackage{stmaryrd}  

\usepackage{tikz-cd}

\usepackage{stackengine}

\newcommand{\cat}[1]{\ensuremath{\mathbf{#1}}}
\newcommand{\catC}{\cat{C}}

\usepackage{tikz,xypic}
\usetikzlibrary{decorations.pathreplacing,decorations.markings,arrows.meta,backgrounds}
\pgfdeclarelayer{edgelayer}
\pgfdeclarelayer{nodelayer}
\pgfsetlayers{background,edgelayer,nodelayer,main}
\tikzstyle{whitedot}=[circle, draw=black, fill=white, inner sep=.4ex]
\tikzstyle{none}=[inner sep=0mm]

\usepackage{tikz,xypic}

\usetikzlibrary{decorations.pathreplacing,decorations.markings,arrows.meta,backgrounds}
\pgfdeclarelayer{edgelayer}
\pgfdeclarelayer{nodelayer}
\pgfsetlayers{background,edgelayer,nodelayer,main}

\tikzstyle{cdot}=[circle, draw=black, fill=black!25, inner sep=.4ex] 
\tikzstyle{bigdot}=[dot, inner sep=0pt]
\tikzstyle{whitedot}=[circle, draw=black, fill=white, inner sep=.4ex]
\tikzstyle{greydot}=[circle, draw=black, fill=black!25, inner sep=.4ex] 
\tikzstyle{blackdot}=[circle, draw=black, fill=black, inner sep=.4ex]
\tikzset{arrow/.style={decoration={
    markings,
    mark=at position #1 with \arrow{>[length=2pt, width=3pt]}},
    postaction=decorate},
    reverse arrow/.style={decoration={
    markings,
    mark=at position #1 with {{\arrow{<[length=2pt, width=3pt]}}}},
    postaction=decorate}
}

\newenvironment{pic}[1][] {\begin{aligned}\begin{tikzpicture}[scale=2.0, font=\tiny,#1]}{\end{tikzpicture}\end{aligned}} 

\newif\ifvflip\pgfkeys{/tikz/vflip/.is if=vflip}
\newif\ifhflip\pgfkeys{/tikz/hflip/.is if=hflip}
\newif\ifhvflip\pgfkeys{/tikz/hvflip/.is if=hvflip}

\newenvironment{picc}[1][]
{\begin{aligned}\begin{tikzpicture}[font=\tiny,#1]}
{\end{tikzpicture}\end{aligned}}

\newlength\minimummorphismwidth
\newlength\stateheight
\newlength\minimumstatewidth
\newlength\connectheight
\tikzset{colour/.initial=white}

\tikzstyle{pure}=[line width=.7pt]

\makeatletter

\pgfdeclareshape{groundd}
{
    \savedanchor\centerpoint
    {
        \pgf@x=0pt
        \pgf@y=0pt
    }
    \anchor{center}{\centerpoint}
    \anchorborder{\centerpoint}

    \anchor{north}
    {
        \pgf@x=0pt
        \pgf@y=0.16\stateheight
    }
    \anchor{south}
    {
        \pgf@x=0pt
        \pgf@y=0pt
    }
    \saveddimen\overallwidth
    {
        \pgfkeysgetvalue{/pgf/minimum width}{\minwidth}
        \pgf@x=\minimumstatewidth
        \ifdim\pgf@x<\minwidth
            \pgf@x=\minwidth
        \fi
    }
    \backgroundpath
    {
        \begin{pgfonlayer}{main} 
        \pgfsetstrokecolor{black}
        \pgfsetlinewidth{1.25pt}
        \ifhflip
            \pgftransformyscale{-1}
        \fi
        \pgftransformscale{0.5}
        \pgfpathmoveto{\pgfpoint{-0.5*\overallwidth}{0pt}}
        \pgfpathlineto{\pgfpoint{0.5*\overallwidth}{0pt}}
        \pgfpathmoveto{\pgfpoint{-0.33*\overallwidth}{0.33*\stateheight}}
        \pgfpathlineto{\pgfpoint{0.33*\overallwidth}{0.33*\stateheight}}
        \pgfpathmoveto{\pgfpoint{-0.16*\overallwidth}{0.66*\stateheight}}
        \pgfpathlineto{\pgfpoint{0.16*\overallwidth}{0.66*\stateheight}}
        \pgfpathmoveto{\pgfpoint{-0.02*\overallwidth}{\stateheight}}
        \pgfpathlineto{\pgfpoint{0.02*\overallwidth}{\stateheight}}
        \pgfusepath{stroke}
        \end{pgfonlayer}
    }
}

\usepackage{tikz,xypic}
\usetikzlibrary{decorations.pathreplacing,decorations.markings,arrows.meta,backgrounds,shapes}
\usetikzlibrary{circuits.ee.IEC}
\pgfdeclarelayer{edgelayer}
\pgfdeclarelayer{nodelayer}
\pgfsetlayers{background,edgelayer,nodelayer,main}
\tikzstyle{none}=[inner sep=0mm]
\tikzstyle{every loop}=[]
\tikzstyle{mark coordinate}=[inner sep=0pt,outer sep=0pt,minimum size=3pt,fill=black,circle]

\tikzset{arrow/.style={decoration={
    markings,
    mark=at position #1 with \arrow{>[length=2pt, width=3pt]}},
    postaction=decorate},
    reverse arrow/.style={decoration={
    markings,
    mark=at position #1 with {{\arrow{<[length=2pt, width=3pt]}}}},
    postaction=decorate}
}

\tikzstyle{upground}=[circuit ee IEC,thick,ground,rotate=90,scale=1.5]
\tikzstyle{upgroundwhite}=[circuit ee IEC,thick,ground,rotate=90,scale=1.5, fill=white]
\tikzstyle{downground}=[circuit ee IEC,thick,ground,rotate=-90,scale=1.5]
\tikzstyle{downgroundnorm}=[circuit ee IEC,thick,ground,rotate=-90,scale=1.5, fill=white]

\newcommand{\mapminh}{5mm} 
\newcommand{\stateminh}{5mm}
\newcommand{\maplw}{0.7pt} 
\newcommand{\stateshift}{-0.2pt}
\newcommand{\effectshift}{-0.2pt}

\tikzstyle{box}=[map]
\tikzstyle{medium box}=[medium map]
\tikzstyle{dot}=[inner sep=0mm,minimum width=2mm,minimum height=2mm,draw,shape=circle]  
\tikzstyle{black dot}=[dot,fill=black]
\tikzstyle{white dot}=[dot,fill=white,,text depth=-0.2mm]
\tikzstyle{grey dot}=[dot,fill=black!25] 

\tikzstyle{corner1}=[box,fill=white, font=\footnotesize] %
\tikzstyle{corner2}=[dot,fill=white, font=\footnotesize] %
\tikzstyle{corner3}=[dot,fill=black!25, font=\footnotesize] %
\tikzstyle{corner4}=[dot,fill=black, font=\footnotesize] %

\tikzstyle{scalar}=[circle,draw,inner sep=2pt, line width=\maplw] 

\usetikzlibrary{shapes.misc, positioning}

\tikzset{stateshape/.style={append after command={
   \pgfextra
        \draw[sharp corners, fill=white, line width = \maplw]%
    (\tikzlastnode.west)%
    [rounded corners=0pt] |- (\tikzlastnode.north)%
    [rounded corners=0pt] -| (\tikzlastnode.east)%
    [rounded corners=5pt] |- (\tikzlastnode.south)%
    [rounded corners=5pt] -| (\tikzlastnode.west);
   \endpgfextra}}}

\tikzset{effectshape/.style={append after command={
   \pgfextra
        \draw[sharp corners, fill=white, line width = \maplw]%
    (\tikzlastnode.west)%
    [rounded corners=0pt] |- (\tikzlastnode.south)%
    [rounded corners=0pt] -| (\tikzlastnode.east)%
    [rounded corners=5pt] |- (\tikzlastnode.north)%
    [rounded corners=5pt] -| (\tikzlastnode.west);
   \endpgfextra}}}

 \tikzstyle{map}=[draw,shape=rectangle, inner sep=2pt,minimum height=\mapminh, minimum width=5mm,fill=white]

\tikzstyle{point}=[stateshape,inner sep=2pt, minimum width=6mm, minimum height=\stateminh, yshift=\stateshift]
\tikzstyle{copoint}=[effectshape,inner sep=.2pt, minimum width=6mm, minimum height=\stateminh, yshift=-\effectshift]

\tikzstyle{wide point}=[point, minimum width=12mm]
\tikzstyle{wide copoint}=[copoint, minimum width=12mm]

\tikzstyle{decomp}=[fill=white,draw,shape=isosceles triangle,shape border rotate=-90,isosceles triangle stretches=true,inner sep=0pt,minimum width=0.75cm,minimum height=4mm,yshift=-0.0mm]

\tikzstyle{decompwide}=[fill=white,draw,shape=isosceles triangle,shape border rotate=-90,isosceles triangle stretches=true,inner sep=0pt,minimum width=1.5cm,minimum height=4mm,yshift=-0.0mm]

\tikzstyle{decompflip}=[fill=white,draw,shape=isosceles triangle,shape border rotate=90,isosceles triangle stretches=true,inner sep=0pt,minimum width=0.75cm,minimum height=4mm,yshift=-0.0mm]

\tikzstyle{decompwideflip}=[fill=white,draw,shape=isosceles triangle,shape border rotate=90,isosceles triangle stretches=true,inner sep=0pt,minimum width=1.5cm,minimum height=4mm,yshift=-0.0mm]

\tikzstyle{medium map} = [map, minimum width = 12mm] 
\tikzstyle{semilarge map} = [map, minimum width = 15mm] 
\tikzstyle{large map} = [map, minimum width = 18mm] 

\tikzstyle{kpoint} =[point]
\tikzstyle{kpointadj} =[copoint]
\tikzstyle{kpointconj}=[dagpointconj] 

\makeatletter
\newcommand{\boxshape}[3]{%
\pgfdeclareshape{#1}{
\inheritsavedanchors[from=rectangle] 
\inheritanchorborder[from=rectangle]
\inheritanchor[from=rectangle]{center}
\inheritanchor[from=rectangle]{north}
\inheritanchor[from=rectangle]{south}
\inheritanchor[from=rectangle]{west}
\inheritanchor[from=rectangle]{east}
\backgroundpath{
\southwest \pgf@xa=\pgf@x \pgf@ya=\pgf@y
\northeast \pgf@xb=\pgf@x \pgf@yb=\pgf@y

\@tempdima=#2
\@tempdimb=#3

\pgfpathmoveto{\pgfpoint{\pgf@xa - 5pt + \@tempdima}{\pgf@ya}}
\pgfpathlineto{\pgfpoint{\pgf@xa - 5pt - \@tempdima}{\pgf@yb}}
\pgfpathlineto{\pgfpoint{\pgf@xb + 5pt + \@tempdimb}{\pgf@yb}}
\pgfpathlineto{\pgfpoint{\pgf@xb + 5pt - \@tempdimb}{\pgf@ya}}
\pgfpathlineto{\pgfpoint{\pgf@xa - 5pt + \@tempdima}{\pgf@ya}}
\pgfpathclose
}
}}

\boxshape{NEbox}{0pt}{3pt} 
\boxshape{SEbox}{0pt}{-3pt}
\boxshape{NWbox}{3pt}{0pt}
\boxshape{SWbox}{-3pt}{0pt}
\boxshape{rec-box}{0pt}{0pt}
\makeatother

\tikzstyle{cloud}=[shape=cloud,draw,minimum width=1.5cm,minimum height=1.5cm]

\tikzstyle{dagmap}=[draw,shape=NEbox,inner sep=2pt,minimum height=\mapminh,fill=white, line width = \maplw] %
\tikzstyle{dashedmap}=[draw,dashed,shape=NEbox,inner sep=2pt,minimum height=\mapminh,fill=white, line width = \maplw]
\tikzstyle{mapdag}=[draw,shape=SEbox,inner sep=2pt,minimum height=\mapminh,fill=white, line width = \maplw]
\tikzstyle{mapadj}=[draw,shape=SEbox,inner sep=2pt,minimum height=\mapminh,fill=white, line width = \maplw]
\tikzstyle{maptrans}=[draw,shape=SWbox,inner sep=2pt,minimum height=\mapminh,fill=white, line width = \maplw]
\tikzstyle{mapconj}=[draw,shape=NWbox,inner sep=2pt,minimum height=\mapminh,fill=white, line width = \maplw]

\tikzstyle{medium dagmap}=[draw,shape=NEbox,inner sep=2pt,minimum height=\mapminh,fill=white,minimum width=7mm, line width = \maplw]
\tikzstyle{semilarge dagmap}=[draw,shape=NEbox,inner sep=2pt,minimum height=\mapminh,fill=white,minimum width=9.5mm, line width = \maplw]
\tikzstyle{large dagmap}=[draw,shape=NEbox,inner sep=2pt,minimum height=\mapminh,fill=white,minimum width=12mm, line width = \maplw]

\makeatletter

\pgfdeclareshape{cornerpoint}{
\inheritsavedanchors[from=rectangle] 
\inheritanchorborder[from=rectangle]
\inheritanchor[from=rectangle]{center}
\inheritanchor[from=rectangle]{north}
\inheritanchor[from=rectangle]{south}
\inheritanchor[from=rectangle]{west}
\inheritanchor[from=rectangle]{east}
\backgroundpath{
\southwest \pgf@xa=\pgf@x \pgf@ya=\pgf@y
\northeast \pgf@xb=\pgf@x \pgf@yb=\pgf@y

\pgfmathsetmacro{\pgf@shorten@left}{\pgfkeysvalueof{/tikz/shorten left}}
\pgfmathsetmacro{\pgf@shorten@right}{\pgfkeysvalueof{/tikz/shorten right}}

\pgfpathmoveto{\pgfpoint{0.5 * (\pgf@xa + \pgf@xb)}{\pgf@ya - 5pt}}
\pgfpathlineto{\pgfpoint{\pgf@xa - 8pt + \pgf@shorten@left}{\pgf@yb - 1.5 * \pgf@shorten@left}}
\pgfpathlineto{\pgfpoint{\pgf@xa - 8pt + \pgf@shorten@left}{\pgf@yb}}
\pgfpathlineto{\pgfpoint{\pgf@xb + 8pt - \pgf@shorten@right}{\pgf@yb}}
\pgfpathlineto{\pgfpoint{\pgf@xb + 8pt - \pgf@shorten@right}{\pgf@yb - 1.5 * \pgf@shorten@right}}
\pgfpathclose
}
}

\pgfdeclareshape{cornercopoint}{
\inheritsavedanchors[from=rectangle] 
\inheritanchorborder[from=rectangle]
\inheritanchor[from=rectangle]{center}
\inheritanchor[from=rectangle]{north}
\inheritanchor[from=rectangle]{south}
\inheritanchor[from=rectangle]{west}
\inheritanchor[from=rectangle]{east}
\backgroundpath{
\southwest \pgf@xa=\pgf@x \pgf@ya=\pgf@y
\northeast \pgf@xb=\pgf@x \pgf@yb=\pgf@y

\pgfmathsetmacro{\pgf@shorten@left}{\pgfkeysvalueof{/tikz/shorten left}}
\pgfmathsetmacro{\pgf@shorten@right}{\pgfkeysvalueof{/tikz/shorten right}}

\pgfpathmoveto{\pgfpoint{0.5 * (\pgf@xa + \pgf@xb)}{\pgf@yb + 5pt}}
\pgfpathlineto{\pgfpoint{\pgf@xa - 8pt + \pgf@shorten@left}{\pgf@ya + 1.5 * \pgf@shorten@left}}
\pgfpathlineto{\pgfpoint{\pgf@xa - 8pt + \pgf@shorten@left}{\pgf@ya}}
\pgfpathlineto{\pgfpoint{\pgf@xb + 8pt - \pgf@shorten@right}{\pgf@ya}}
\pgfpathlineto{\pgfpoint{\pgf@xb + 8pt - \pgf@shorten@right}{\pgf@ya + 1.5 * \pgf@shorten@right}}
\pgfpathclose
}
}

\makeatother

\pgfkeyssetvalue{/tikz/shorten left}{0pt}
\pgfkeyssetvalue{/tikz/shorten right}{0pt}

\tikzstyle{dagpoint common}=[draw,fill=white,inner sep=1pt, line width = \maplw, minimum height = 4mm, yshift=1.2pt] 
\tikzstyle{dagpoint sc}=[shape=cornerpoint,dagpoint common]
\tikzstyle{dagpoint adjoint sc}=[shape=cornercopoint,dagpoint common]
\tikzstyle{dagpoint}=[shape=cornerpoint,shorten left=4pt,dagpoint common]
\tikzstyle{dagpointadj}=[shape=cornercopoint,shorten left=5pt,dagpoint common]
\tikzstyle{dagpointconj}=[shape=cornerpoint,shorten right=5pt,dagpoint common]
\tikzstyle{dagpointtrans}=[shape=cornercopoint,shorten right=5pt,dagpoint common]
\tikzstyle{dagpointsymm}=[shape=cornerpoint,shorten left=5pt,shorten right=5pt,dagpoint common]

\tikzstyle{widedagpoint}=[dagpoint, minimum width=1 cm, inner sep=2pt]
\tikzstyle{widedagpointadj}=[dagpointadj, minimum width=1 cm, inner sep=2pt]

\tikzstyle{every picture}=[baseline=-0.25em,scale=0.5]
\tikzstyle{label}=[font=\footnotesize,text height=1ex, text depth=0.15ex]

\usetikzlibrary[shapes]

\tikzset{
sidetriangle/.style = {regular polygon, regular polygon sides = 3, aspect = 1, shape border rotate = 90, draw, inner sep = 0, minimum width = 1.2cm}
}

\tikzset{
isoc/.style = {shape=isosceles triangle, shape border rotate = 180, isosceles triangle stretches = true, minimum width = 1.2cm, minimum height= 1.5cm, inner sep = 0.3}}

\tikzset{
coarse/.style = {shape = circle, fill = white, draw, inner sep = 0, minimum width =0.125cm}
}
\tikzset{
coarsesymbol/.style = {shape = circle, fill = white, inner sep = -0.7, minimum width = 0.125cm}
}

\tikzstyle{sidetriangle2}=[sidetriangle, minimum width = 2cm, fill=white]
\tikzstyle{sideisocsmall}]=[style=isoc, minimum width = 1cm, minimum height = 0.8cm, draw, fill=white, font=\Large]
\tikzstyle{sideisoc}]=[style=isoc, minimum width = 2cm, draw, fill=white, font=\Large]
\tikzstyle{sideisocmid}]=[style=isoc, minimum width = 2.5cm, draw, fill=white, font=\Large]
\tikzstyle{sideisocmedium}]=[style=isoc, minimum width = 3cm, draw, fill=white, font=\Large]

\tikzstyle{label}=[font=\footnotesize,text height=1ex, text depth=0.15ex]

\tikzstyle{box}=[map]
\tikzstyle{medium box}=[medium map]
\tikzstyle{dot}=[inner sep=0mm,minimum width=2mm,minimum height=2mm,draw,shape=circle]  
\tikzstyle{black dot}=[dot,fill=black]
\tikzstyle{white dot}=[dot,fill=white,,text depth=-0.2mm]
\tikzstyle{grey dot}=[dot,fill=black!25] 

\tikzstyle{corner1}=[box,fill=white, font=\footnotesize] %
\tikzstyle{corner2}=[dot,fill=white, font=\footnotesize] %
\tikzstyle{corner3}=[dot,fill=black!25, font=\footnotesize] %
\tikzstyle{corner4}=[dot,fill=black, font=\footnotesize] %

\tikzstyle{scalar}=[circle,draw,inner sep=2pt, line width=\maplw] 

\usetikzlibrary{shapes.misc, positioning}

\tikzset{stateshape/.style={append after command={
   \pgfextra
        \draw[sharp corners, fill=white, line width = \maplw]%
    (\tikzlastnode.west)%
    [rounded corners=0pt] |- (\tikzlastnode.north)%
    [rounded corners=0pt] -| (\tikzlastnode.east)%
    [rounded corners=5pt] |- (\tikzlastnode.south)%
    [rounded corners=5pt] -| (\tikzlastnode.west);
   \endpgfextra}}}

\tikzset{effectshape/.style={append after command={
   \pgfextra
        \draw[sharp corners, fill=white, line width = \maplw]%
    (\tikzlastnode.west)%
    [rounded corners=0pt] |- (\tikzlastnode.south)%
    [rounded corners=0pt] -| (\tikzlastnode.east)%
    [rounded corners=5pt] |- (\tikzlastnode.north)%
    [rounded corners=5pt] -| (\tikzlastnode.west);
   \endpgfextra}}}

\tikzstyle{point}=[stateshape,inner sep=2pt, minimum width=6mm, minimum height=\stateminh, yshift=\stateshift]
\tikzstyle{copoint}=[effectshape,inner sep=.2pt, minimum width=6mm, minimum height=\stateminh, yshift=-\effectshift]

\tikzstyle{wide point}=[point, minimum width=12mm]
\tikzstyle{wide copoint}=[copoint, minimum width=12mm]

\begin{document}

\maketitle

\begin{abstract}
    In this report we present a new model of concepts, based on the framework of variational autoencoders, which is designed to have attractive properties such as factored conceptual domains, and at the same time be learnable from data. The model is inspired by, and closely related to, the $\beta$-VAE model of concepts, but is designed to be more closely connected with language, so that the names of concepts form part of the graphical model. We provide evidence that our model---which we call the \emph{Conceptual VAE}---is able to learn interpretable conceptual representations from simple images of coloured shapes together with the corresponding concept labels. We also show how the model can be used as a concept classifier, and how it can be adapted to learn from fewer labels per instance. Finally, we formally relate our model to G\"ardenfors' theory of conceptual spaces, showing how the Gaussians we use to represent concepts can be formalised in terms of ``fuzzy concepts" in such a space.  
\end{abstract}

\section{Introduction}

The philosophical and psychological study of concepts has a long tradition in philosophy, linguistics, psychology and cognitive science \cite{murphy_concepts,conceptual_mind}. There is also a large body of work on formal, mathematical models of concepts \cite{Ganter2016}.
More recently, AI researchers have begun to investigate how concepts can be learned from raw perceptual data \cite{beta-vae}, in the hope that an artificial agent that has induced conceptual representations from its  environment will be able to reason and act more effectively in that environment, similar to how humans use concepts \shortcite{lake_thinking_machines}.

This report is inspired by the work of G\"{a}rdenfors, who has developed a theory of concepts based on the idea that concepts form convex regions in some geometric space \cite{gardenfors,gardenfors2014}. Our conceptual representations also live in a geometric (latent) space, but unlike G\"{a}rdenfors' theory they are probabilistic in nature. Having ``fuzzy" probabilistic representations not only provides a natural mechanism for dealing with the vagueness inherent in the human conceptual system, but also allows us to draw on the  toolkit from machine learning to provide effective learning mechanisms. Here we follow \citeA{beta-vae} in using the framework of Variational Autoencoders (VAEs) \shortcite{rezende14,kingma14} to learn  conceptual representations from simple images. One contribution of this report is to define a new type of VAE---which we call the \emph{Conceptual VAE}---which explicitly links a conceptual representation with the word(s) used to refer to that concept (e.g. \emph{red circle}). The concepts themselves are multivariate Gaussians living in a factored conceptual space.

We use the Spriteworld software \shortcite{spriteworld19} to generate simple images consisting of coloured shapes of certain sizes in certain positions, meaning our conceptual spaces contain four conceptual \emph{domains} \cite{gardenfors}: \textsc{colour}, \textsc{size}, \textsc{shape} and \textsc{position}.\footnote{We use small caps for domains and italics for concepts and their labels.} These domains provide the factors of the conceptual space.
The main question we investigate is a representational learning one: can the Conceptual VAE induce factored representations in a latent conceptual space which neatly separates the individual concepts, and under what conditions? Here we demonstrate that, if the system is provided with supervision regarding the domains, and provided with the corresponding four labels for each training instance (e.g. (\emph{blue, small, circle, top}\/)), then the VAE can learn Gaussians which faithfully represent the colour spectrum, for example. Moreover, we extend the Conceptual VAE---using a Gaussian mixture model for the conceptual ``prior" representations---so that it is able to learn from fewer labels for each training instance, including just one (e.g. \emph{blue}). Finally, we show how the Conceptual VAE naturally provides a concept classifier, in the form of the encoder, which predicts a Gaussian for an image that can be compared with the induced conceptual representations using the KL divergence. The concept with the smallest KL relative to the encoding of that image can then be selected.

As well as defining the Conceptual VAE, and describing experimental work using simple images, another contribution of this report is to provide a formalisation of our model which is consistent with G\"{a}rdenfors' framework. Our aim is to present a precisely defined model---so that it is  clear, from a mathematical point of view, what concepts are in our model---which we can also implement and train in practice.

The rest of the report is structured as follows. Section~\ref{sec:conceptual_vae} describes our conceptual VAE model, by first explaining the standard VAE; then showing how a concept label can be introduced via the conditional VAE; and finally showing how to modify these model architectures to get the Conceptual VAE. This section also explains how the Conceptual VAE is trained. Section~\ref{sec:classifier} shows how the encoder from the Conceptual VAE naturally provides a concept classifier, and Section~\ref{sec:fewer_labels} describes how the model can be adapted to deal with training sets where only a subset of the concept labels is provided as supervision per instance.

Section~\ref{sec:expts} describes the experiments we have performed, first describing the dataset of coloured shapes from which the model induces conceptual representations (\ref{sec:shapes}), and then describing the neural architecture of the model (\ref{sec:neural_arch}). Section~\ref{sec:clustering} demonstrates how the means and variances predicted by the encoder neatly cluster along the relevant dimensions. Section~\ref{sec:classification} presents some results from the concept classifier. Section~\ref{sec:concept_order} investigates how the conceptual representations are ordered along a dimension, including some experiments on an extended-colour dataset, with all the colours of the rainbow. And Section~\ref{sec:any_expts} gives some results for the Gaussian mixture model which is trained using fewer labels per instance.

Section~\ref{sec:formal} provides a mathematical formalisation of conceptual spaces (including ``fuzzy" concepts), based on \citeA{tull2021categorical}, and relates it to our  probabilistic models. 
Section~\ref{sec:related_work} provides a brief survey of some existing work which is closely related to ours (including the $\beta$-VAE of \citeA{beta-vae}), and Section~\ref{sec:further_work} concludes the report, including some ideas for future work.

\section{VAEs for Concept Modelling}
\label{sec:conceptual_vae}

The Variational Autoencoder (VAE) \shortcite{kingma14,rezende14} provides a framework for the generative modeling of data, where the data potentially lives in some high-dimensional space. It uses the power of neural networks to act as arbitrary function approximators to capture complex dependencies in the data (e.g. between the pixels in an image). The VAE uses a latent space $\mathbf{Z}$ which acts as a bottleneck, compressing the high-dimensional data into a lower dimensional space. The question we  investigate in this report is whether the VAE model can be adapted so that $\mathbf{Z}$ has desirable properties from a conceptual space perspective, such as interpretable dimensions which contain neatly separated, labelled concepts from individual domains. In the next section we describe the standard VAE model, which we refer to as a \emph{vanilla VAE}, before describing how to adapt it in order to incorporate labelled concepts. 

\subsection{The Vanilla VAE}

\begin{figure}
    \centering
        \[
        \begin{tikzpicture}
	\begin{pgfonlayer}{nodelayer}
		\node [style=seen] (0) at (2.5, -3) {$\mathbf{x}$};
		\node [style=unseen] (1) at (2.5, 0) {$\mathbf{z}$};
		\node [style=none] (2) at (1.25, -4.5) {};
		\node [style=none] (3) at (1.25, 1.25) {};
		\node [style=none] (4) at (3.75, 1.25) {};
		\node [style=none] (5) at (3.75, -4.5) {};
		\node [style=none] (6) at (3.75, 0.75) {};
		\node [style=none] (7) at (4.5, -3) {$\theta$};
		\node [style=none] (8) at (4.25, -3) {};
		\node [style=seen] (9) at (18.75, -3) {$\mathbf{x}$};
		\node [style=unseen] (10) at (18.75, 0) {$\mathbf{z}$};
		\node [style=seen] (11) at (18.75, 3) {$\mathbf{c}$};
		\node [style=seen] (12) at (11, 0) {$\mathbf{c}$};
		\node [style=none] (13) at (0.75, 0) {};
		\node [style=none] (14) at (0.5, 0) {$\phi$};
		\node [style=none] (15) at (3.25, -4) {$N$};
		\node [style=none] (16) at (20.5, 0) {};
		\node [style=none] (17) at (20.75, 0) {$\psi$};
		\node [style=none] (18) at (17, 0) {};
		\node [style=none] (19) at (16.75, 0) {$\phi$};
		\node [style=none] (20) at (20.75, -3) {$\theta$};
		\node [style=none] (21) at (20.5, -3) {};
		\node [style=seen] (22) at (9.5, -3) {$\mathbf{x}$};
		\node [style=unseen] (23) at (9.5, 0) {$\mathbf{z}$};
		\node [style=none] (24) at (8.25, -4.5) {};
		\node [style=none] (25) at (8.25, 1.25) {};
		\node [style=none] (26) at (12, 1.25) {};
		\node [style=none] (27) at (12, -4.5) {};
		\node [style=none] (28) at (12, 0.75) {};
		\node [style=none] (29) at (12.75, -3) {$\theta$};
		\node [style=none] (30) at (12.5, -3) {};
		\node [style=none] (31) at (7.75, 0) {};
		\node [style=none] (32) at (7.5, 0) {$\phi$};
		\node [style=none] (33) at (11.5, -4) {$N$};
		\node [style=none] (34) at (17.5, -4.5) {};
		\node [style=none] (35) at (17.5, 3.75) {};
		\node [style=none] (36) at (20, 3.75) {};
		\node [style=none] (37) at (20, -4.5) {};
		\node [style=none] (38) at (20, 3.25) {};
		\node [style=none] (39) at (19.5, -4) {$N$};
	\end{pgfonlayer}
	\begin{pgfonlayer}{edgelayer}
		\draw [style=full] (1) to (0);
		\draw (5.center) to (2.center);
		\draw (2.center) to (3.center);
		\draw (6.center) to (5.center);
		\draw (6.center) to (4.center);
		\draw (4.center) to (3.center);
		\draw [style=full] (8.center) to (0);
		\draw [style=full] (10) to (9);
		\draw [style=full] (11) to (10);
		\draw [style=full] (16.center) to (10);
		\draw [style=full] (21.center) to (9);
		\draw [style=full] (23) to (22);
		\draw (27.center) to (24.center);
		\draw (24.center) to (25.center);
		\draw (28.center) to (27.center);
		\draw (28.center) to (26.center);
		\draw (26.center) to (25.center);
		\draw [style=full] (30.center) to (22);
		\draw [style=densely dashed, ->] (31.center) to (23);
		\draw [style=full] (12) to (22);
		\draw (37.center) to (34.center);
		\draw (34.center) to (35.center);
		\draw (38.center) to (37.center);
		\draw (38.center) to (36.center);
		\draw (36.center) to (35.center);
		\draw [style=densely dashed, ->] (13.center) to (1);
		\draw [style=densely dashed, ->, bend left] (0) to (1);
		\draw [style=densely dashed, ->, bend left] (9) to (10);
		\draw [style=densely dashed, ->, bend left] (22) to (23);
		\draw [style=densely dashed, ->] (18.center) to (10);
	\end{pgfonlayer}
\end{tikzpicture}
        \] 
    \caption{Graphical models for the VAE (left), conditional VAE (centre) and the Conceptual VAE (right). Grey nodes represent observed variables and white nodes hidden variables.}
    \label{fig:vae}
\end{figure}
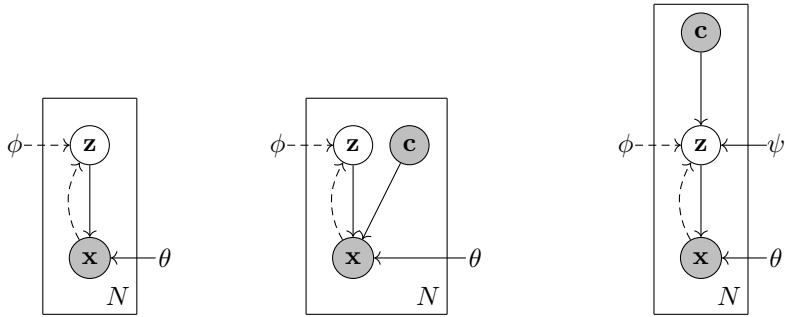

Figure~\ref{fig:vae} (left) shows the graphical model for the VAE, which has a particularly simple structure. In terms of the generative story, which is represented by the solid arrows in the plate diagram, first a point $\z$ in the latent space $\Z$ is sampled according to the prior $p(\z)$, and then a data point $\x$ is generated according to the likelihood $p_\theta(\x|\z)$. This process is assumed to have been repeated $N$ times to generate some dataset $\mathbf{X}$, and the goal of learning is to find a reasonable set of model parameters under that assumption, given a suitable parametrisation. The dashed arrows denote the variational approximation $q_\phi(\mathbf{z}|\mathbf{x})$ to the intractable posterior $p_\theta(\mathbf{z}|\mathbf{x})$. A main innovation of the VAE framework is to instantiate $p$ and $q$ using neural networks, parametrised by $\theta$ and $\phi$ respectively.

The prior is assumed to be a centered isotropic multivariate Gaussian $p(\mathbf{z}) = \mathcal{N}(\mathbf{z};\mathbf{0},\mathbf{1})$ \cite{kingma14}. The approximate posterior $q_\phi(\mathbf{z}|\mathbf{x})$ is also assumed to be a multivariate Gaussian with a diagonal covariance matrix, but with means and variances  predicted by a neural network with learnable parameters $\phi$. In our case, since $\mathbf{X}$ is a dataset of images, $q_\phi$ will be instantiated by a convolutional neural network (CNN), which is referred to as the \emph{encoder}. Similarly, $p_\theta$ will be instantiated by a de-convolutional neural network (de-CNN), and referred to as the \emph{decoder}. The encoder is used during training; intuitively its job is to learn which parts of the latent space are likely to have generated a training instance, so that the optimisation of the loss function can focus only on that part of the space. 

The function that is optimised during training is the RHS of the following equation \cite{doersch}:
\begin{equation}
    \log p(\x) - \mathcal{D}(q(\z),p(\z|\x)) = \mathbb{E}_{\z\sim q(\z)}[\log p(\x|\z)] - \mathcal{D}(q(\z),p(\z))
    \label{eqn:elbo}
\end{equation} 
\noindent
where $\mathcal{D}$ is the KL divergence. Note that the LHS contains the expression we would really like to optimise: the (log-)likelihood $\log p(\x)$. The idea is that if we can instantiate $q$ using a powerful neural network, then during training the KL between $q(\z)$ and the true posterior $p(\z|\x)$ can be driven close to zero, in which case we'd be optimising the likelihood. Note also that, since the KL on the LHS is positive, the equation provides a lower bound on the likelihood, known as the \emph{evidence lower bound} (ELBO). Equation~\ref{eqn:elbo} holds for any $q$, and so it makes sense to have $q$ depend on $\x$, in which case the equation becomes:
\begin{equation}
\log p(\x) - \mathcal{D}(q(\z|\x),p(\z|\x)) =
    \mathbb{E}_{\z\sim q(\z|\x)}[\log p(\x|\z)] - \mathcal{D}(q(\z|\x),p(\z))
    \label{eqn:elbo_zx}
\end{equation}

The advantage of this formulation is that the RHS can be maximised using gradient-based optimisation techniques. Since the KL on the RHS is between two multivariate Gaussians, there is an analytical expression for calculating this quantity. An estimate of the expectation can be obtained using numerical methods, in particular Monte Carlo sampling; however, there is a technical difficulty with the use of sampling during gradient-based optimisation which is that the sampling step destroys any gradients. Another innovation in the VAE framework is the so-called \emph{reparametrisation trick}, in which the samples are drawn from the unit normal as part of an input layer to the network, but then rescaled according to the means and variances of $q$ \cite{kingma14,rezende14}. That way all the computation steps within the network are continuous and gradients can pass all the way through \cite{doersch}.

Perhaps the easiest way to understand the optimisation of~(\ref{eqn:elbo_zx}) is to consider the steps taken during training. During each iteration of training, given an instance $\x$, first the means and variances of $q(\z|\x)$ are predicted using the encoder, and the KL between $q$ and the prior $p$ is calculated. Then, a sample $\z_s$ is taken from $q$ and $\log p(\x|\z_s)$ is calculated, using the decoder, which is the (negative) cost of reconstructing the image using that sample.\footnote{More than one sample could be used to get a better Monte Carlo estimate of the expectation, but one sample is typically sufficient in practice.} If we think of the negative of the RHS of~(\ref{eqn:elbo_zx}) as a loss, then 
the loss function has two parts: the KL loss and the reconstruction loss. The job of the encoder is to predict a region in $\Z$ which is likely to have generated $\x$ (resulting in a low reconstruction loss from the decoder) but at the same time one which is not too far from the unit Gaussian (giving a low KL loss with respect to the prior).

Are the latent representations induced by a VAE in any way \emph{conceptual}? First, note that there is no pressure within the model to induce the sorts of factored representations in which the dimensions of $\Z$ correspond to conceptual domains. \citeA{beta-vae}, discussed below in Section~\ref{sec:related_work}, attempts to address this problem by introducing a weighting factor on the KL loss. Second, there is currently no mechanism in the model which allows concepts to be referred to using their names (e.g. \emph{blue square}). The next section is an attempt to address this latter problem.

\subsection{The Conditional VAE}

One feature that we would like in the model is an explicit representation of the words or symbols that are used to refer to a concept (which we'll call the concept \emph{label}). The link between language and concepts is a feature of the human conceptual system, and the idea that language provides a ``window into the mind" has become something of a truism in the cognitive sciences \cite{pinker}.\footnote{Exactly how language and concepts are related is a contentious issue, but for one particular view on the  connection see \citeA{evans}.} Given the tight relationship between words and concepts, we expect that language data may provide a useful signal for learning conceptual representations themselves. For example, the classes of words used to refer to concepts in different domains could be induced from patterns of word usage in large text corpora.\footnote{There is a large literature in NLP on using patterns in text to learn hypernym-, or ISA-, hierarchies, following \citeA{hearst}.} In this report we will assume the domains are known in advance, and the connection between language and conceptual representations is not a question we explore in any detail, but we would still like a model in which the induced conceptual representations have an accompanying label, so that we can answer questions such as ``what is the concept for \emph{red}?".

The obvious way to include the concept label in the model is as an explicit random variable $\con$. Figure~\ref{fig:vae} (centre) shows one way to do that, using the \emph{conditional VAE} \cite{doersch}. Here the label acts as an additional input into the decoder, so that when the decoder generates a  data instance $\x$, it does so conditioned on $\con$ as well as a point from the latent space $\z$: $p_\theta(\x|\z,\con)$.
The ELBO equation~(\ref{eqn:elbo_zx}) now takes the following form \cite{doersch}:
\begin{equation}
\log p(\x|\con) - \mathcal{D}(q(\z|\x,\con),p(\z|\x,\con)) =
    \mathbb{E}_{\z\sim q(\z|\x,\con)}[\log p(\x|\z,\con)] - \mathcal{D}(q(\z|\x,\con),p(\z))
    \label{eqn:elbo_cond}
\end{equation}
\noindent
Note that the encoder $q_\phi$ now takes the label $\con$ as an additional input, and similarly for the decoder $p_\theta$. The prior $p(\z)$ acts independently of $\con$ and so remains as the unit normal $\mathcal{N}(\z;\mathbf{0},\mathbf{1})$.

In this work we assume that $\con$ is factored in terms of the conceptual domains, so that for an instance $\x$ each of the domains has a corresponding atomic label, and hence $\con = (\czero,\cone,\ctwo,\cthree)$. An example concept label would be (\emph{red, small, square, centre}). In terms of the CNN encoder and decoder, there are now four additional inputs, each one a one-hot encoding of an atomic concept label from the corresponding domain vocabulary (colours, sizes, shapes and positions).

The conditional VAE has provided us with a way of incorporating concept labels into the model; however, there is a problem with this proposal, which is that there is no explicit representation of a concept (beyond its symbolic label). If we were to ask the question,  ``what is the representation for \emph{red}", for example, the conditional VAE cannot provide an answer. This is problematic from a theoretical modelling point of view, but also potentially problematic when attempting to furnish an  agent with conceptual representations (since what representations would we use?). The model described in the next section provides an answer to the question.

\subsection{The Conceptual VAE}

The key to the conceptual VAE is to introduce a new random variable for a concept label, $\con$, as for the conditional VAE, but unlike that model introduce it at the very top of the graphical model (Figure~\ref{fig:vae}; right). In terms of the generative story, first a concept label $\con$ is generated, and then a point $\z$ in the latent conceptual space is generated, \emph{conditioned on} $\con$; after that the generative story is the same as for the vanilla VAE: an instance $\x$ is generated conditioned on $\z$. In this work we assume a uniform prior over the concept labels (more specifically a uniform prior over the atomic labels corresponding to each conceptual domain $\con_i$), and $\con$ can effectively be thought of as a fixed input to the model, as provided by the data.   

How do we model $p(\z|\con)$? As before we use multivariate Gaussians with diagonal covariance matrices, but now the means and variances are \emph{learnable parameters} $\psi$. We will sometimes refer to $p_\psi(\z|\con)$ for a given concept $\con$ as a \emph{conceptual ``prior"} (since these Gaussians replace the unit normal prior in the vanilla VAE), as well as $\con$'s learned representation.\footnote{The scare quotes on \emph{prior} are intended to emphasise that these are distributions that are learned from the data.} Since $\con$ is factored, each $\con_i$ has its own (univariate) Gaussian distribution; for example, \emph{red} will have its own mean and variance which define a Gaussian on the dimension corresponding to the \textsc{colour} domain. It is this Gaussian which provides the anwser to the question ``what is the conceptual representation for \emph{red}?". 

The ELBO equation now takes the following form:
\begin{equation}
\log p(\x|\con) - \mathcal{D}(q(\z|\x),p(\z|\x,\con)) =
    \mathbb{E}_{\z\sim q(\z|\x)}[\log p(\x|\z)] - \mathcal{D}(q(\z|\x),p(\z|\con))
    \label{eqn:elbo_con}
\end{equation}
\noindent
Note some differences to the RHS of the ELBO equation for the conditional VAE (\ref{eqn:elbo_cond}). First, the KL is between the encoder distribution $q_\phi$ and the conceptual prior $p_\psi(\z|\con)$; hence rather than trying to fit the encoder distributions to a unit normal, now the objective is to fit each encoder distribution $q_\phi(\z|\x)$ to the corresponding representation for the $\con$ labelling $\x$ (both of which are learned). Second, the decoder term in the reconstruction loss, $\log p_\theta(\x|\z)$, no longer depends on $\con$, since $\x$ is independent of $\con$ given $\z$ (because of the structure of the graphical model in Figure~\ref{fig:vae} (right)). Finally, we have chosen to use $q(\z|\x)$ rather than $q(\z|\x,\con)$ as the encoder distribution, for two reasons. The first reason is that, for the instances in our data, the possible $\con_i$'s on a particular dimension for a given $i$ are mutually exclusive (e.g. an image cannot be both red and green), and so $\x$ provides all the information the model needs to infer $\z$.\footnote{This may not be true if we had overlapping concepts such as \emph{red} and \emph{dark-red}. How to deal with such cases is left for future work.} The second reason is that having the encoder predict $q(\z|\x)$ rather than $q(\z|\x,\con)$ leads to a particularly neat form for the concept classifier based on the KL, described in Section~\ref{sec:classifier}.

How is this model trained, and what are the pressures that lead to conceptual representations being learned? For a training instance $\x$ labelled with a concept $\con$, the training proceeds as before for the vanilla VAE: the encoder predicts a Gaussian $q(\z|\x)$; this is sampled from (using the reparametrisation trick) to give a sample $\z_s$; and $-\log p(\x|\z_s)$ is calculated to give the reconstruction loss. The key difference is in the calculation of the KL loss. Suppose that $\con$ = (\emph{green, medium, triangle, bottom}). The KL is calculated for each dimension, relative to the Gaussian for the particular atomic label for that dimension. For example, for the \textsc{colour} domain (dimension 0), the KL would be between $q_\phi(\z_0|\x)$ and $p_\psi(\z_0|\mbox{\emph{green}})$. So note that the supervision regarding the domains is provided here in the calculation of the KL.
Unlike the vanilla VAE, the conceptual ``priors" depend on the learned parameters $\psi$, which are the means and variances of the individual (univariate) Gaussians. We expect these learned means and variances to result in a neat separation along a dimension, since this will make it easier for the model to fit $q$ to the conceptual representations, leading to a lower KL. And indeed this is what happens, as shown in the various analyses in Section~\ref{sec:expts} below.

\subsubsection{A Concept Classifier}
\label{sec:classifier}

In Section~\ref{sec:expts} we will perform some qualitative evaluation, showing how the atomic concepts (such as the individual colours) neatly separate along a single dimension. It would also be useful to have a more quantitative evaluation demonstrating that the model is behaving as we expect. An obvious candidate is classification: given a red shape, for example, can the model use its conceptual representation for \emph{red} to correctly identify the shape's colour? Note that the classification task itself, from a computer vision perspective, is trivial, and one that we would expect a well-trained CNN to solve. The classification task is being used here as a test of whether the induced conceptual representations can be employed in a useful way.

From a probabilistic perspective, the goal is to find the most probable concept $\con'$ given an input image $\x$:
\begin{eqnarray}
    \con' & = &\arg\max_\con p(\con|\x) \label{bayes} \\
    \label{bayes2}
          & = & \arg\max_\con p(\x|\con)\\ \label{bayes3}
          & \approx & \arg\max_\con\; - \mathcal{D}(q(\z|\x),p(\z|\con)) + \mbox{recon\_loss} \\
          \label{bayes4}
          & = & \arg\max_\con\;  - \mathcal{D}(q(\z|\x),p(\z|\con))
\end{eqnarray}
Line (\ref{bayes2}) follows from (\ref{bayes}) because of the assumed uniform prior over concepts, and we use the ELBO from (\ref{eqn:elbo_con}) as an approximation to the likelihood in going from (\ref{bayes2}) to (\ref{bayes3}). The reconstruction loss is independent of $\con$ and so we end up with the satisfying form of the classifier in (\ref{bayes4}), in which the most likely concept for an input $\x$ is the one with the smallest KL relative to the encoding of $\x$, as provided by $q$. Since the model is factored in terms of the conceptual domains, the $\arg\max$ can be carried out independently for each dimension; e.g. the most probable colour for a shape can be determined efficiently by calculating the KL for all the atomic concepts on the colour dimension only.

Note that we could easily build a concept classifier using the conditional VAE, but it would not take the satisfying form just described for the Conceptual VAE. First, there are no explicit conceptual representations in the conditional VAE; and second the decoder is not applied independently of $\con$ given $\z$ in the conditional VAE, and so $p(\x|\z,\con)$ would need to be used when calculating the score for concept $\con$.

\subsubsection{Fewer Labels per Instance}
\label{sec:fewer_labels}

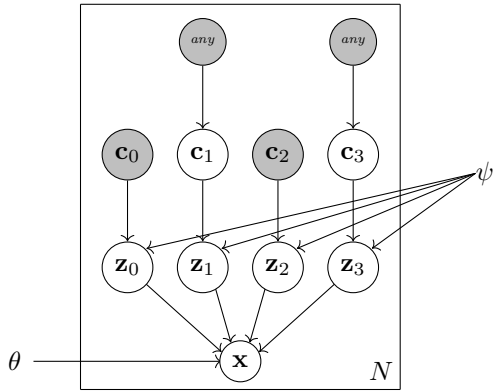
\begin{figure}
    \centering
    \[
    \begin{tikzpicture}
	\begin{pgfonlayer}{nodelayer}
		\node [style=unseen] (0) at (9, -3) {$\z_0$};
		\node [style=seen] (1) at (9, 0) {$\con_0$};
		\node [style=none] (3) at (18.5, -0.5) {$\psi$};
		\node [style=none] (4) at (7.75, 4) {};
		\node [style=none] (5) at (16.25, 4) {};
		\node [style=none] (6) at (7.75, -6.25) {};
		\node [style=none] (7) at (16.25, -6.25) {};
		\node [style=none] (8) at (18.25, -0.5) {};
		\node [style=none] (10) at (9.5, -2.5) {};
		\node [style=unseen] (11) at (11, -3) {$\z_1$};
		\node [style=unseen] (12) at (11, 0) {$\con_1$};
		\node [style=seen] (13) at (11, 3) {$\mbox{\tiny \emph{any}}$};
		\node [style=unseen] (14) at (13, -3) {$\z_2$};
		\node [style=seen] (15) at (13, 0) {$\con_2$};
		\node [style=unseen] (17) at (15, -3) {$\z_3$};
		\node [style=unseen] (18) at (15, 0) {$\con_3$};
		\node [style=seen] (19) at (15, 3) {$\mbox{\tiny \emph{any}}$};
		\node [style=none] (20) at (18.25, -0.5) {};
		\node [style=none] (21) at (11.5, -2.5) {};
		\node [style=none] (22) at (18.25, -0.5) {};
		\node [style=none] (24) at (13.5, -2.5) {};
		\node [style=none] (25) at (18.25, -0.5) {};
		\node [style=none] (26) at (15.5, -2.5) {};
		\node [style=unseen] (27) at (12, -5.5) {$\x$};
		\node [style=none] (28) at (9, -3) {};
		\node [style=none] (29) at (11.5, -5.25) {};
		\node [style=none] (30) at (11, -3) {};
		\node [style=none] (32) at (11.25, -3.25) {};
		\node [style=none] (33) at (11.75, -5) {};
		\node [style=none] (34) at (12.75, -3.25) {};
		\node [style=none] (35) at (12.25, -5) {};
		\node [style=none] (36) at (15, -3) {};
		\node [style=none] (37) at (12.5, -5.25) {};
		\node [style=none] (38) at (6, -5.5) {$\theta$};
		\node [style=none] (39) at (6.5, -5.5) {};
		\node [style=none] (40) at (11.5, -5.5) {};
		\node [style=none] (41) at (15.75, -5.75) {$N$};
	\end{pgfonlayer}
	\begin{pgfonlayer}{edgelayer}
		\draw [style=full] (1) to (0);
		\draw (4.center) to (5.center);
		\draw (7.center) to (5.center);
		\draw (6.center) to (4.center);
		\draw (6.center) to (7.center);
		\draw [style=full] (8.center) to (10.center);
		\draw [style=full] (12) to (11);
		\draw [style=full] (13) to (12);
		\draw [style=full] (15) to (14);
		\draw [style=full] (18) to (17);
		\draw [style=full] (19) to (18);
		\draw [style=full] (20.center) to (21.center);
		\draw [style=full] (22.center) to (24.center);
		\draw [style=full] (25.center) to (26.center);
		\draw [style=full] (28.center) to (29.center);
		\draw [style=full] (32.center) to (33.center);
		\draw [style=full] (34.center) to (35.center);
		\draw [style=full] (36.center) to (37.center);
		\draw [style=full] (39.center) to (40.center);
	\end{pgfonlayer}
\end{tikzpicture}
    \]
    \caption{Graphical model for the Conceptual VAE with the \emph{any} label, when $\con_1$ and $\con_3$ are generated from \emph{any}.}
    \label{fig:vae_any}
\end{figure}

One weakness of the model so far is that we have been assuming that the label $\con$ for each data instance $\x$ contains atomic labels for all four domains: \textsc{colour, size, shape, position}. But what if a data instance comes with just one or two atomic labels? Such flexibility could be useful when considering more realistic datasets, which may contain instances which are labelled with only a subset of the full set of atomic concepts.

Figure~\ref{fig:vae_any} shows how to incorporate an additional label---the \emph{any} label---into the model to allow this flexibility. The idea is that any missing labels will be replaced with the \emph{any} label. The example is for the case when \emph{any} applies to the second and fourth latent dimensions, e.g. (\emph{green, any, circle, any}) which would denote a green circle of unspecified size and position. Since the atomic concepts are assumed to be uniformly distributed for a particular dimension, the graphical model in Figure~\ref{fig:vae_any} represents a Gaussian mixture model with equal probabilities assigned to each component of the mixture (so $1/3$ in the case where there are three atomic labels for each domain, as we have below in our main dataset).

One feature of this model is that, when calculating the loss, only the calculation of the KL is changed; the reconstruction loss is calculated as before. One difficulty is that there is no analytical solution for the calculation of the KL between Gaussian mixtures, and so we need to use a numerical solution. Here we use a Monte Carlo estimate of the KL, as we do for the reconstruction loss (but with more samples). We find that this approach works well for our setup.

\section{Experiments}
\label{sec:expts}

In this section we describe the dataset used in our experiments, the basic architecture used for the neural networks, followed by some qualitative and quantitative analysis of our model.

\subsection{The Shapes Dataset}
\label{sec:shapes}

We use the Spriteworld software \cite{spriteworld19} to generate simple images consisting of coloured shapes of particular sizes in particular positions in a 2D box. For the main dataset, there are three shapes: \{\emph{square, triangle, circle}\}; three colours: \{\emph{red, green, blue}\}; three sizes: \{\emph{small, medium, large}\}; and three positions: \{\emph{bottom, centre, top}\}. The \textsc{colour} attribute here refers to the hue, with the saturation and brightness varied randomly.
The position is relative to the vertical dimension, and the horizontal position is fixed to the centre. The background colour is always the same. 

\begin{figure}
\centering
    \includegraphics[]{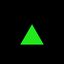}
    \includegraphics[]{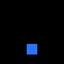}
    \includegraphics[]{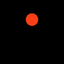}
    \includegraphics[]{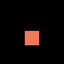}
    \includegraphics[]{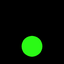}
    \caption{Example shapes, from left to right: (\emph{green, large, triangle, centre}); (\emph{blue, small, square, bottom}); 
    (\emph{red, medium, circle, top});
    (\emph{red, medium, square, centre});
    (\emph{green, large, circle, bottom}).}
    \label{fig:shapes}
\end{figure}

Figure~\ref{fig:shapes} shows some example shapes.
The examples nicely demonstrate the vagueness and variety inherent in the underlying concepts generating the data. For example, the red square is towards the orange end of redness, and the medium-sized circle is close in size to the small blue square. 
Appendix~\ref{sec:app_shapes} contains the parameters used in the Spriteworld software to generate the main dataset. The parameters give a range of values for
each of the atomic labels (other than those for \textsc{shape} which are discrete). The appropriate ranges are sampled from uniformly, given a tuple of 4 uniformly sampled input labels, to give the particular values used to generate an instance. We ran the sampler to generate a training set of 3,000 instances, and development  and test sets with 300 instances each.

\subsection{The Neural Networks}
\label{sec:neural_arch}

The encoder, which takes an image $\x$ as input, is instantiated as a CNN, with 4 convolutional layers followed by a fully-connected layer. A final layer predicts the means and variances of the multivariate Gaussian $q_\phi(\z|\x)$. The ReLU activation function is used throughout (except for the final layer).
The decoder, which takes a latent point $\z$ as input, is instantiated as a de-CNN, with essentially the mirrored architecture of the encoder. The reconstruction loss we use on the decoder for predicting the pixel values in an image $\x$ is the MSE loss.\footnote{We also tried the cross-entropy loss, following \citeA{doersch}, but found that the MSE loss gave better results in practice.}
Appendix~\ref{sec:app_neural_nets} contains more details of the neural architectures used in our experiments, including the various hyper-parameter choices.

The implementation was in Tensorflow. The full set of parameters to be learned is $\theta \cup \phi \cup \psi$, where $\theta$ is the set of parameters in the encoder, $\phi$ the parameters in the decoder, and $\psi$ the means and variances for the conceptual representations (12 each for the main dataset). The training was run for 200 epochs (unless stated otherwise), with a batch size of 32, and the Adam optimizer was used. 

Finally, we added 2 ``slack" dimensions to the latent space $\Z$, in addition to the 4 dimensions for the conceptual domains. These slack dimensions are intended to capture any remaining variability in the images, beyond that contained in the concepts themselves. Since our images are relatively simple, it is possible that the slack dimensions are not needed here, but we included them (unless stated otherwise) in anticipation of scaling up the model to more complex images.

\subsection{Clustering Effects for the Encoder}
\label{sec:clustering}

\begin{figure}
    \hspace*{-0.5cm}\includegraphics[scale=0.22]{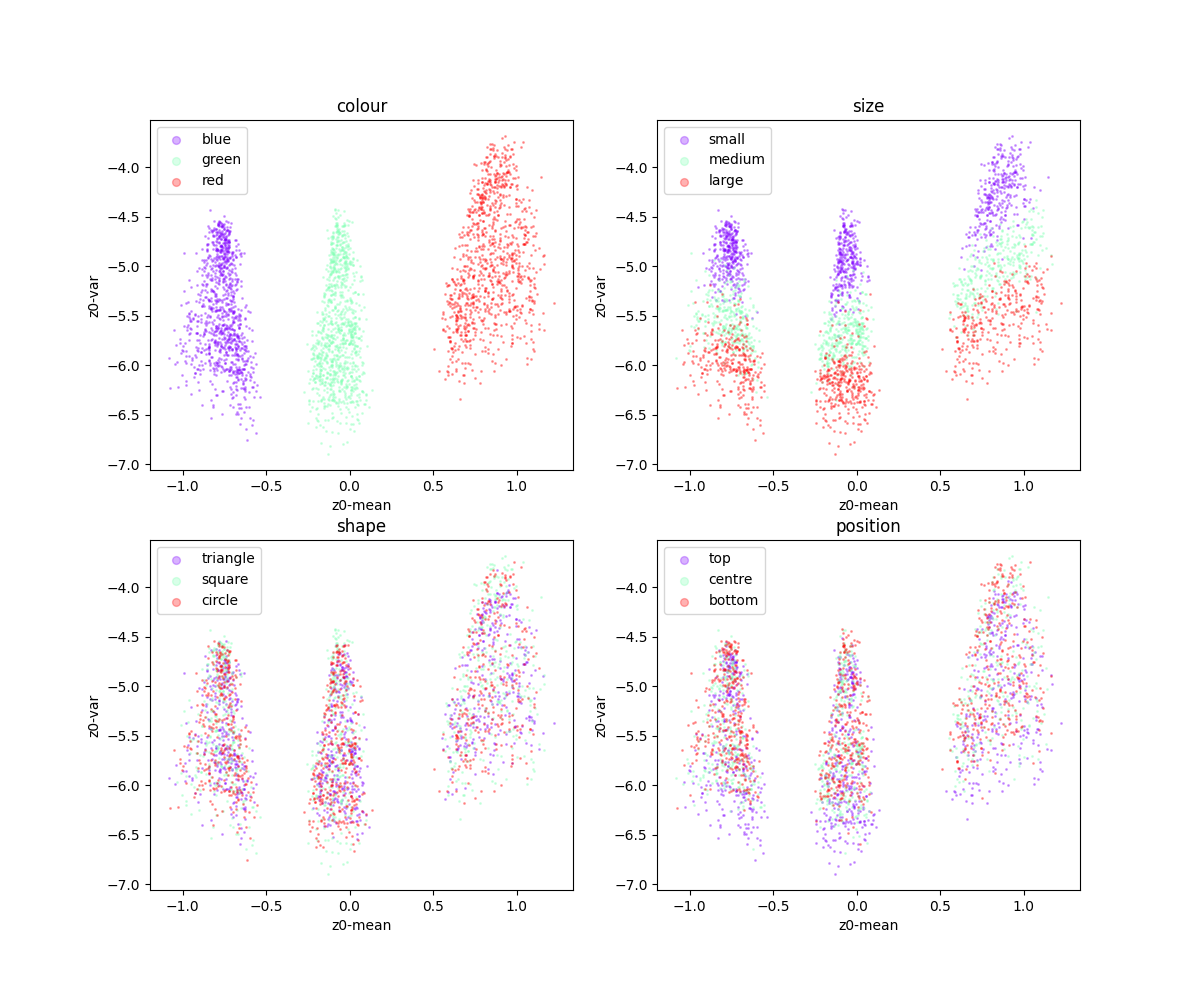}
    \includegraphics[scale=0.22]{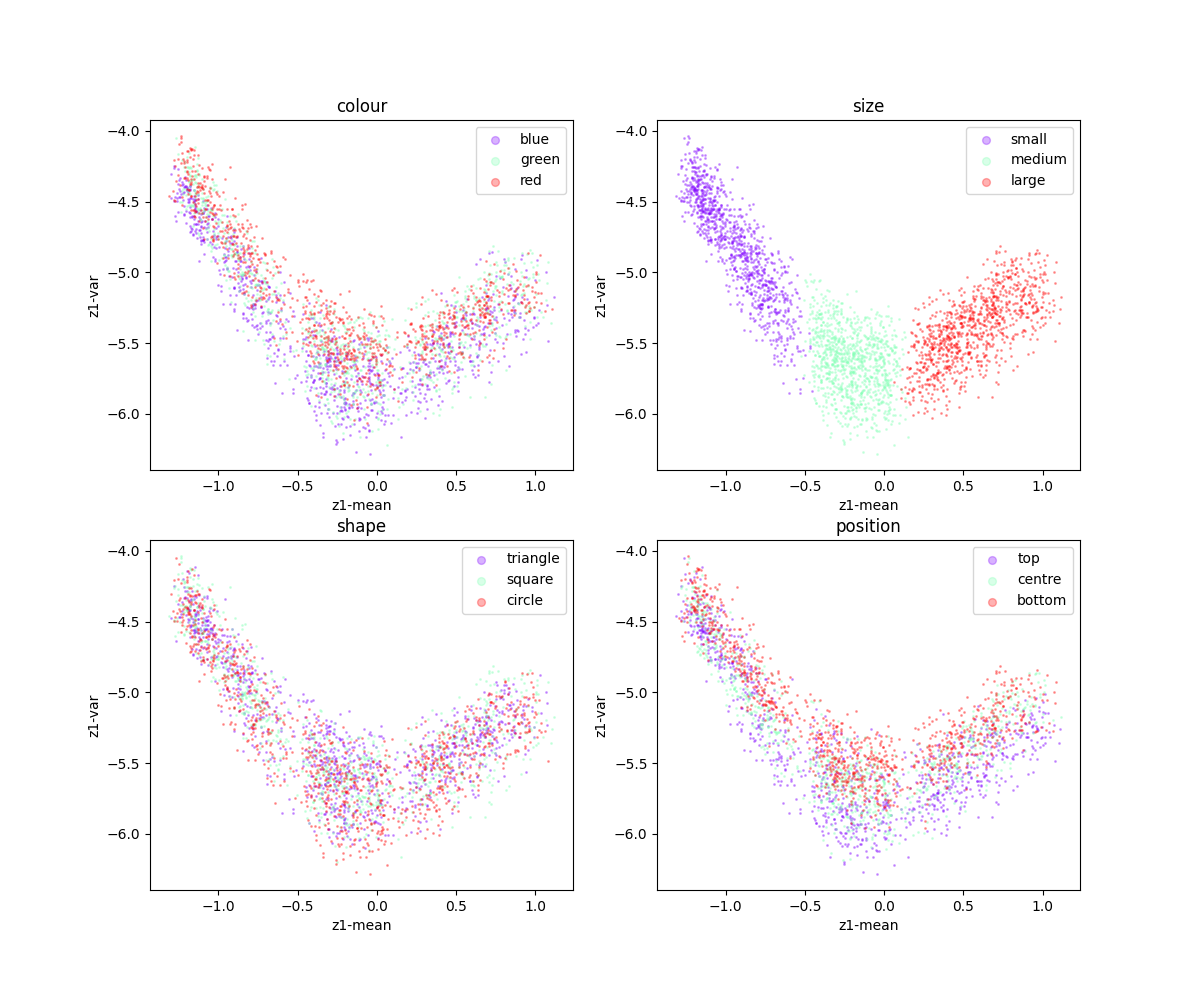}
    \hspace*{-0.5cm}\includegraphics[scale=0.22]{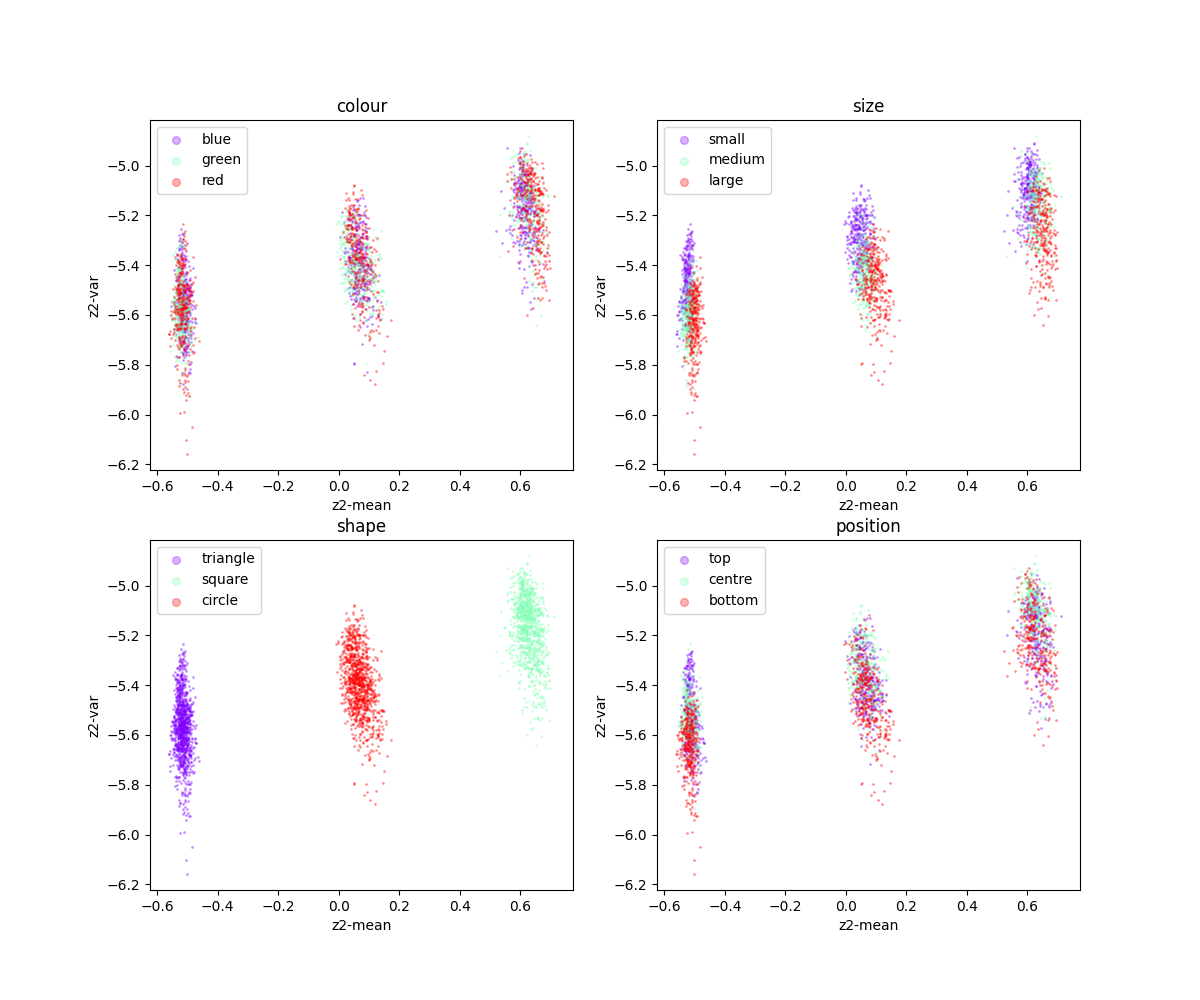}
    \includegraphics[scale=0.22]{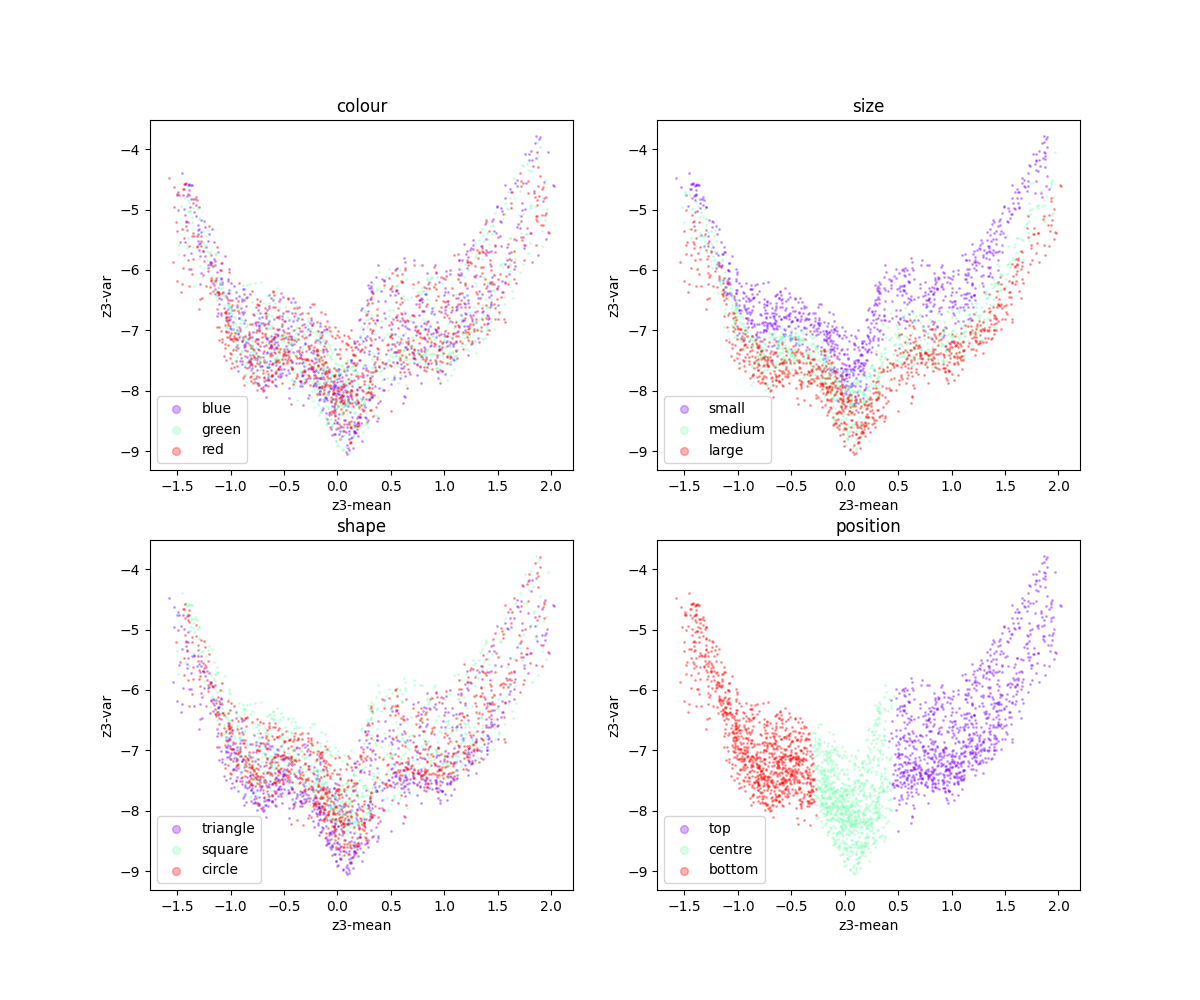}
    \hspace*{-0.5cm}\includegraphics[scale=0.22]{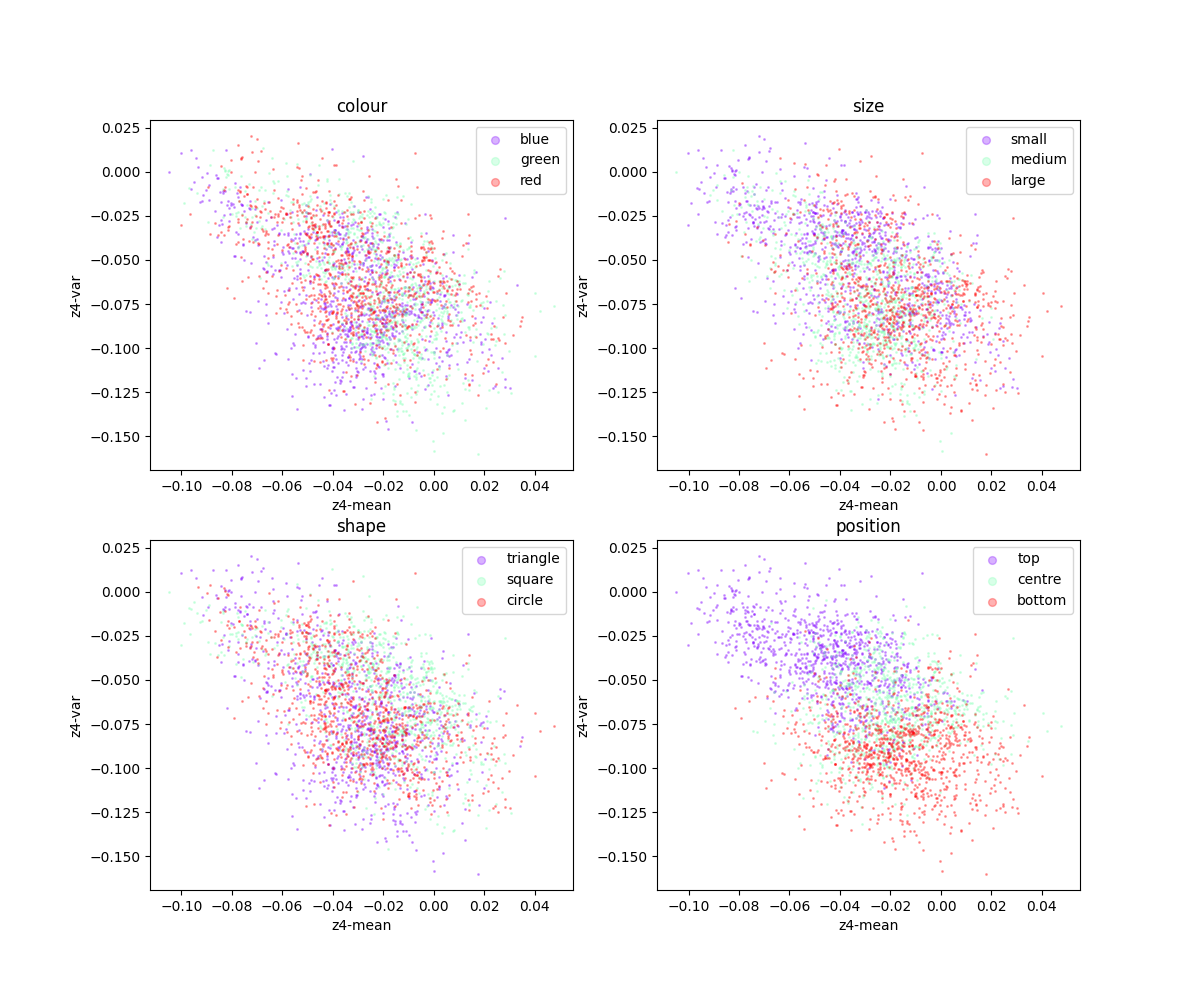}
    \includegraphics[scale=0.22]{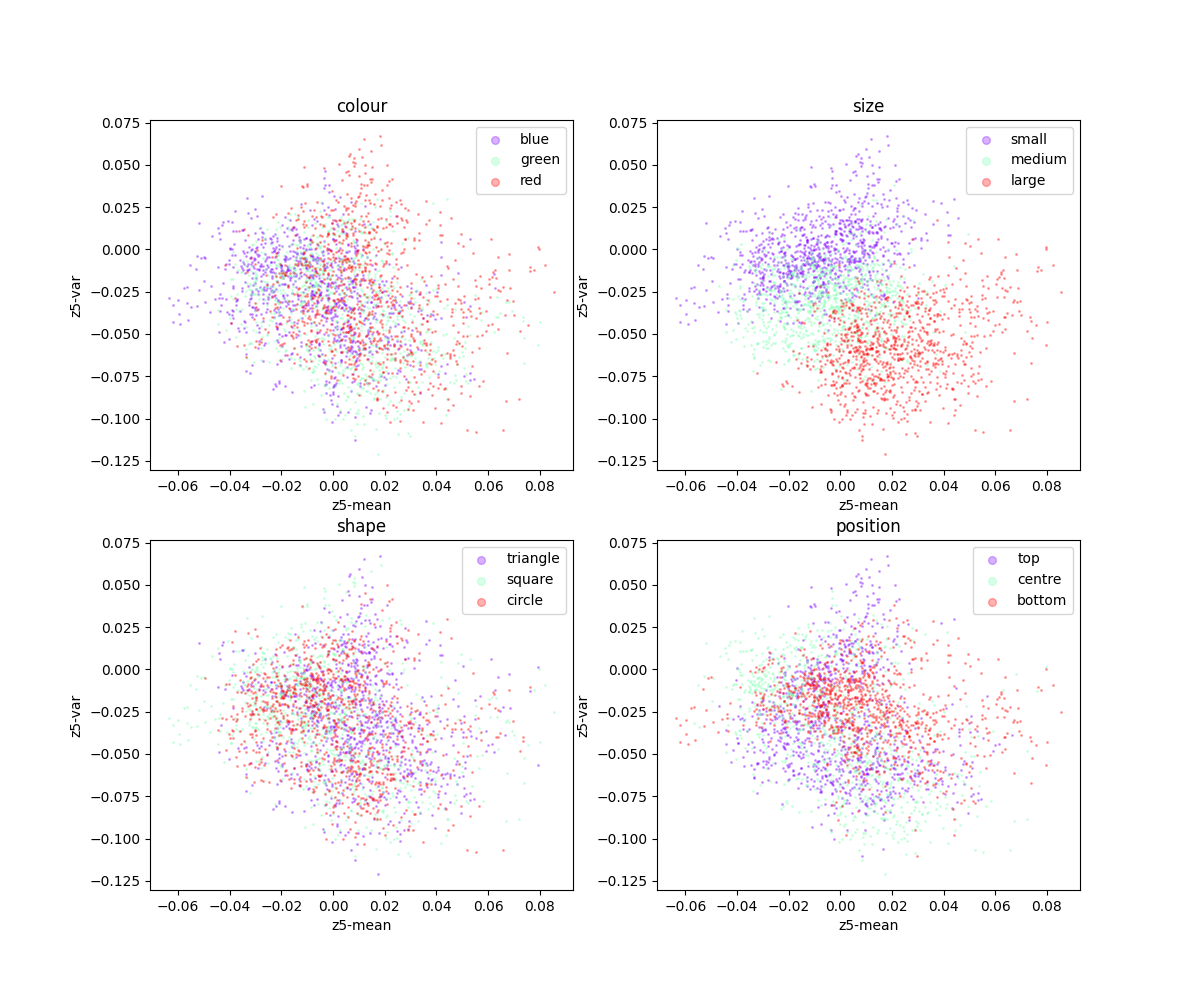}
    \caption{The means and log-variances for each dimension predicted by the encoder, for a set of instances, colour-coded by the atomic concept labels; means on the x-axis, log-variances on the y-axis. Colour-coding, from top-left clockwise: \textsc{colour, size, position}, slack-dim-2, slack-dim-1, \textsc{shape}.}
    \label{fig:clustering}
\end{figure}

Figure~\ref{fig:clustering} shows the means and log-variances predicted by the encoder for each dimension, for a set of instances, with the colour-coding indicating the atomic concept labels from the different domains. For example, in the set of 4 plots in the top-left, the means and log-variances for dimension 0 are plotted; and in the top-left of those 4 plots, each point is colour-coded with the colour of the corresponding instance. What this plot shows is the neat separation for the means along the \textsc{colour} dimension, for each of the 3 colours. The other 3 plots contain the same set of points, but colour-coded with atomic labels from the remaining domains of \textsc{size}, \textsc{shape} and \textsc{position}. With the 3 remaining plots we expect to see no discerning pattern, since we would like the first dimension to encode \textsc{colour} only.

The plots were created using the model evaluated in Section~\ref{sec:classification} below, which performed well in the classification task on the development data. The instances were taken from the training data.\footnote{The same patterns were observed on the development data. We used the training data since this gives denser plots.} The plots in the top-right are for dimension 1 (corresponding to \textsc{size}), and again we obtain a neat separation for the means, when colour-coded with the size of the instance, with instances labelled \emph{medium} sitting in the middle.\footnote{Anecdotally we observe that instances labelled \emph{medium} tend to be placed between those labelled \emph{small} and \emph{large}. Sec.~\ref{sec:concept_order} investigates the ordering of instances for the \textsc{colour} domain.} The middle-right plots are for dimension 3 (\textsc{position}), and again we see a neat separation of the means with instances labelled \emph{centre} sitting between those labelled \emph{top} and \emph{bottom}. The middle-left plots are for dimension 2 (\textsc{shape}). Here we see a clear separation with the predicted means occupying a short range, which reflects the discrete nature of these concepts. Finally, the bottom 2 plots are for the slack dimensions, and here we expect to see no discernible pattern, since these dimensions are not intended to capture any information about the conceptual domains.

\begin{table}[t]
    \centering
    \begin{tabular}{@{}llll@{}}
    \toprule
    domain                             & concept         & mean & log-var \\* \midrule
    \multirow{3}{*}{\textsc{colour}}   & \emph{blue}     & -0.77 & -4.23  \\* \cmidrule(l){2-4} 
                                       & \emph{green}    & -0.08 & -4.31  \\* \cmidrule(l){2-4} 
                                       & \emph{red}      & ~0.83 & -3.56  \\* \midrule
    \multirow{3}{*}{\textsc{size}}     & \emph{small}    & -0.93 & -3.01  \\* \cmidrule(l){2-4} 
                                       & \emph{medium}   & -0.20 & -3.62  \\* \cmidrule(l){2-4} 
                                       & \emph{large}    & ~0.56 & -2.82  \\* \midrule
    \multirow{3}{*}{\textsc{shape}}    & \emph{triangle} & -0.49 & -5.31  \\* \cmidrule(l){2-4} 
                                       & \emph{square}   & ~0.64 & -4.91  \\* \cmidrule(l){2-4} 
                                       & \emph{circle}   & ~0.09 & -5.08  \\* \midrule
    \multirow{3}{*}{\textsc{position}} & \emph{top}      & ~1.07 & -1.82  \\* \cmidrule(l){2-4} 
                                       & \emph{centre}   & ~0.04 & -3.33  \\* \cmidrule(l){2-4} 
                                       & \emph{bottom}   & -0.81 & -2.37  \\* \bottomrule
    \end{tabular}
    \caption{Learned Gaussians for the concept representations.}
    \label{tab:learned_priors}
\end{table}

Note that there does appear to be some information encoded for \textsc{size} in the slack dimension 5 (bottom-right). Also, there are some patterns displaying ``vertically" in the plots (e.g. dimension 0 when colour-coded by size), indicating that some of the conceptual information is being encoded in the variances of the corresponding Gaussians, rather than the means. But overall the patterns displayed in Figure~\ref{fig:clustering} are largely as anticipated. 

Finally, the values of the learned conceptual ``priors" are shown in Table~\ref{tab:learned_priors}. We present this particular set of values to emphasise the fact that an atomic concept is represented by two real numbers -- the mean and variance of its Gaussian. Note also that the means and log-variances predicted by the encoder in Figure~\ref{fig:clustering} are consistent with the values of the concept representations; for example, the means for \emph{top, centre} and \emph{bottom} of 1.07, 0.04 and -0.81, respectively, fit closely the 3 clusters shown in the centre-right plot. This is expected because the training encourages the means and variances predicted by the encoder for an instance to be close to the representation of that instance's concept (through the KL part of the loss).

\subsection{Classifier Results}
\label{sec:classification}

\begin{table}[t]
    \centering
    \begin{tabular}{@{}l|rr@{}}
    \toprule
    & dev & test \\
    \hline
       \textsc{colour}  & 1.00 & 1.00 \\
     \textsc{size} &  0.98 & 0.99 \\
     \textsc{shape} &  1.00 & 1.00 \\
     \textsc{position} & 0.99 & 0.97\\
     \bottomrule
    \end{tabular}
    \caption{Classifier accuracy per domain.}
    \label{tab:classifier_results}
\end{table}

Table~\ref{tab:classifier_results} shows the classifier accuracy per domain, on the development and test data, for a model chosen according to its performance on the development data.\footnote{The training is relatively robust for this model on this dataset, and a large proportion of the randomly initialised models give good accuracies.} As a reminder, the classifier operates on each dimension/domain independently and chooses the atomic concept whose conceptual prior has the lowest KL with that predicted by the encoder (Section~\ref{sec:classifier}). 
We expect that the model does not always give 100\% accuracy for some domains because of the vagueness inherent in some of the instances (which would be difficult for humans to classify correctly; see Section~\ref{sec:shapes}).

\begin{figure}
\centering
        \includegraphics[scale=0.32]{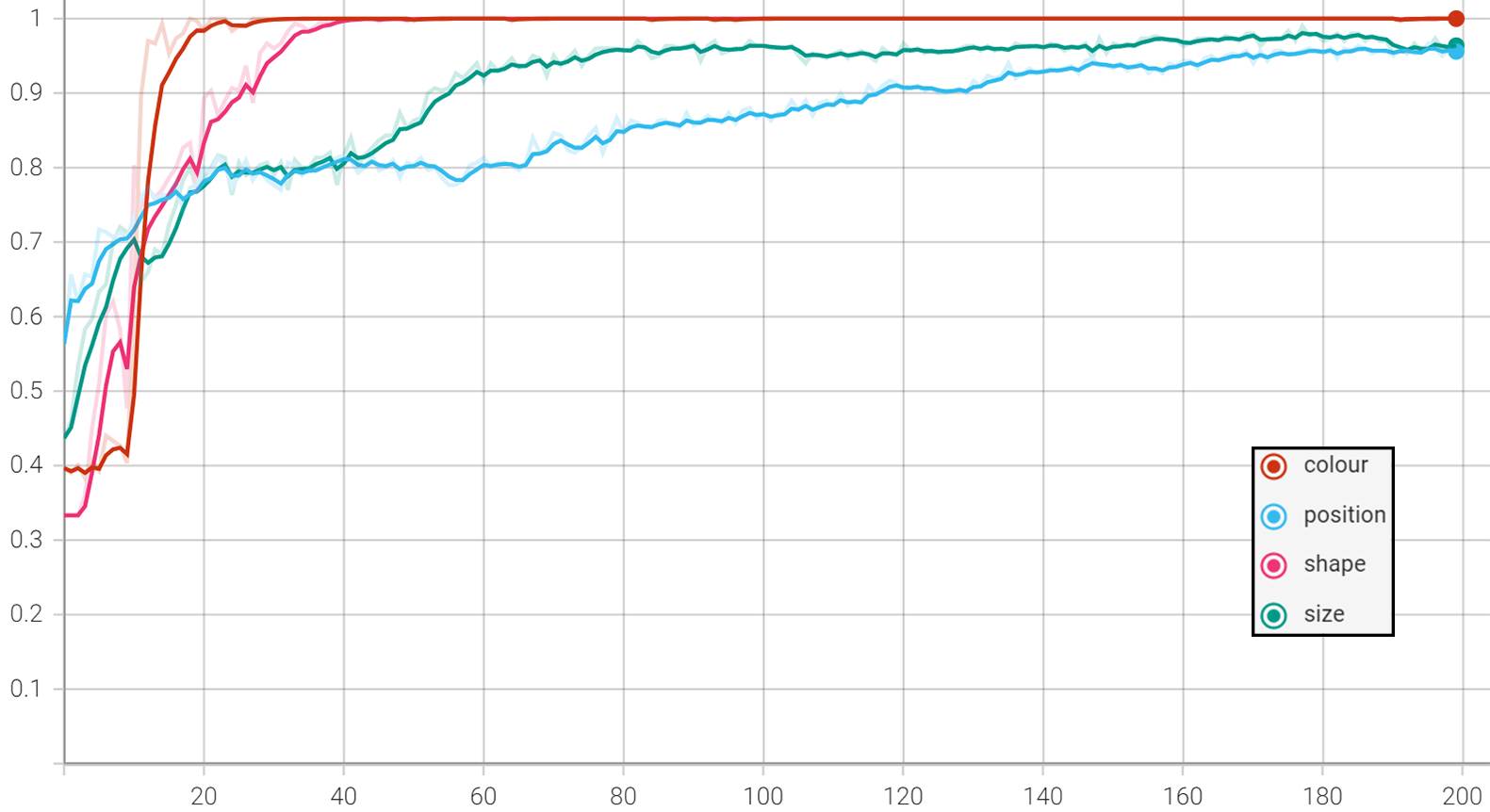}
    \caption{Learning curve showing classifier accuracy on the development data for each domain during training.}
    \label{fig:learning_curve}
\end{figure}

Figure~\ref{fig:learning_curve} is a learning curve showing how the classifier accuracy improves during training, for each domain. The \textsc{colour} domain is learned relatively quickly, with \textsc{position} and \textsc{size} taking longer, which may reflect the higher levels of vagueness for the latter two concepts (for this particular dataset).


\subsection{Concept Ordering}
\label{sec:concept_order}

\begin{figure}
    \centering
    \includegraphics[width=\textwidth]{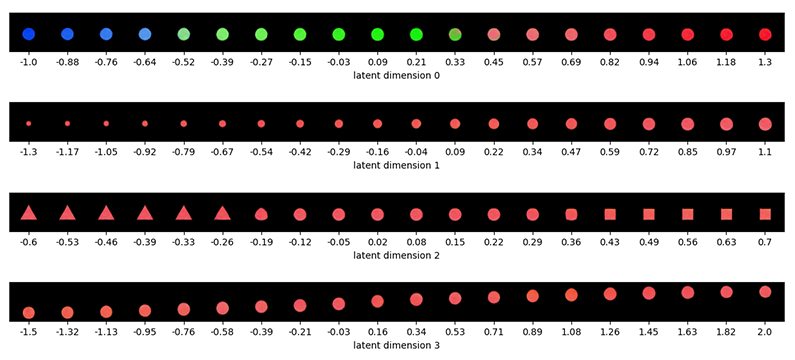}
    \caption{Traversals along each latent dimension for a large red circle in the centre.}
    \label{fig:traversal_example_1}
\end{figure}

\begin{figure}
    \centering
    \includegraphics[width=\textwidth]{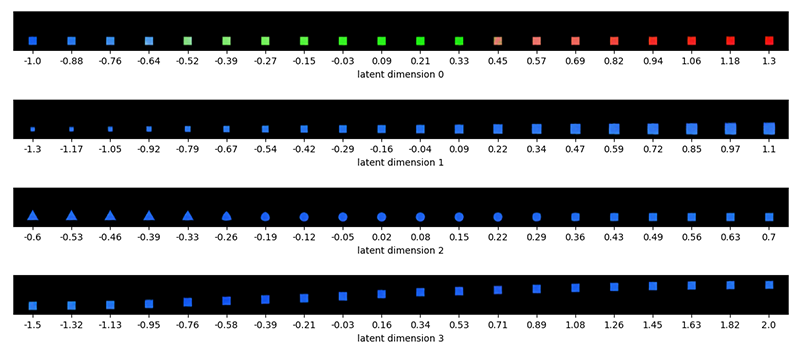}
    \caption{Traversals along each latent dimension for a medium-sized blue square at the bottom.}
    \label{fig:traversal_example_2}
\end{figure}

Figures~\ref{fig:traversal_example_1} and~\ref{fig:traversal_example_2} provide a further qualitative demonstration of how the conceptual domains are neatly represented on each dimension. In Figure~\ref{fig:traversal_example_1}, an instance of a large red circle in the centre is passed through the encoder, giving a mean for each of the 4 dimensions. Then, the mean value is systematically varied for one of the dimensions only (through regular increases and decreases), keeping the other 3 fixed. All resulting combinations of the 4 mean values are then input to the decoder, giving the images in the figure.\footnote{The idea of plotting transitions along a dimension is taken from \citeA{beta-vae}.} 

What the transitions clearly demonstrate is not only how one latent dimension encodes just one domain, but also how the concepts smoothly vary along one dimension. Note how dimension 2 encodes a shape somewhere between a \emph{triangle} and a \emph{circle}, and  also a  shape somewhere between a \emph{circle} and a \emph{square}. Dimension 1 shows a smooth transition from \emph{small} to \emph{medium} to \emph{large}, and dimension 3 from \emph{bottom} to \emph{center} to \emph{top}.
Figure~\ref{fig:traversal_example_2} shows the same traversals but starting with a medium-sized blue square at the bottom.

The analysis so far has not only demonstrated a neat separation of the concept representations along the relevant dimension, but also suggested that the model may respect a natural ordering of the concepts, e.g. placing \emph{medium} between \emph{small} and \emph{large} for the \textsc{size} dimension, and \emph{centre} between \emph{top} and \emph{bottom} for  \textsc{position}. Since the means for the concept representations are initialised randomly (Appendix~\ref{sec:app_neural_nets}), any ordering effect must be due to an inductive bias in the model, as well as being contingent on properties of the data (e.g. continuity across the range of relevant values, for the non-discrete concepts).

In order to investigate these ordering effects further, we created a new dataset 
which contains all the colours of the rainbow, with the same shapes, sizes and positions. Appendix~\ref{sec:ext_shapes} contains the parameters used in the Spriteworld software to generate the extended dataset with more colours. The continuous ranges now cover a much larger proportion of the range of possible values between 0 and 1, with the occasional gap (e.g. between \emph{green} and \emph{blue}). The training data again consisted of 3,000 randomly generated instances, with a development set of 300 instances.

\begin{figure}
    \hspace*{-0.5cm}\includegraphics[scale=0.22]{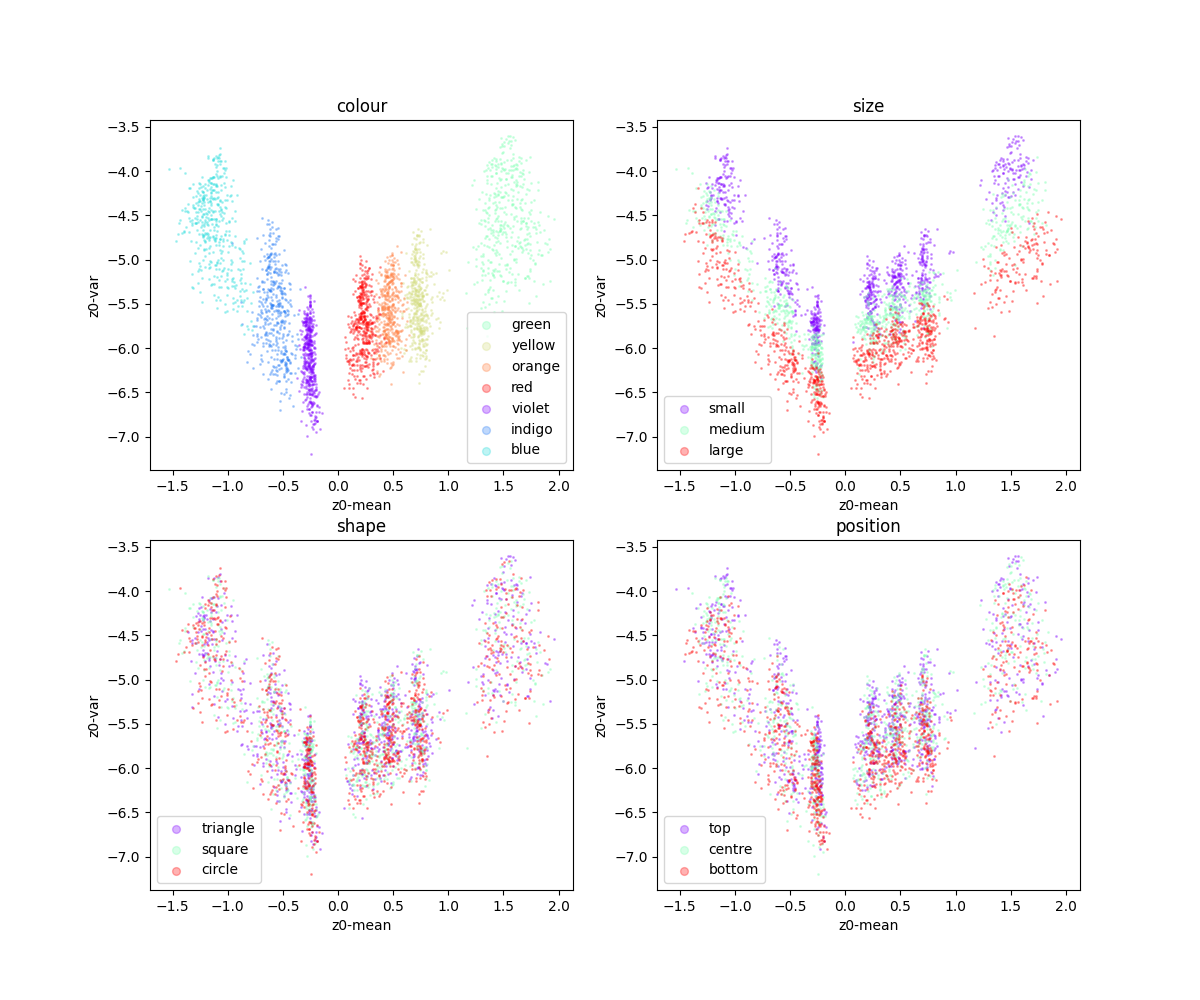}
    \includegraphics[scale=0.22]{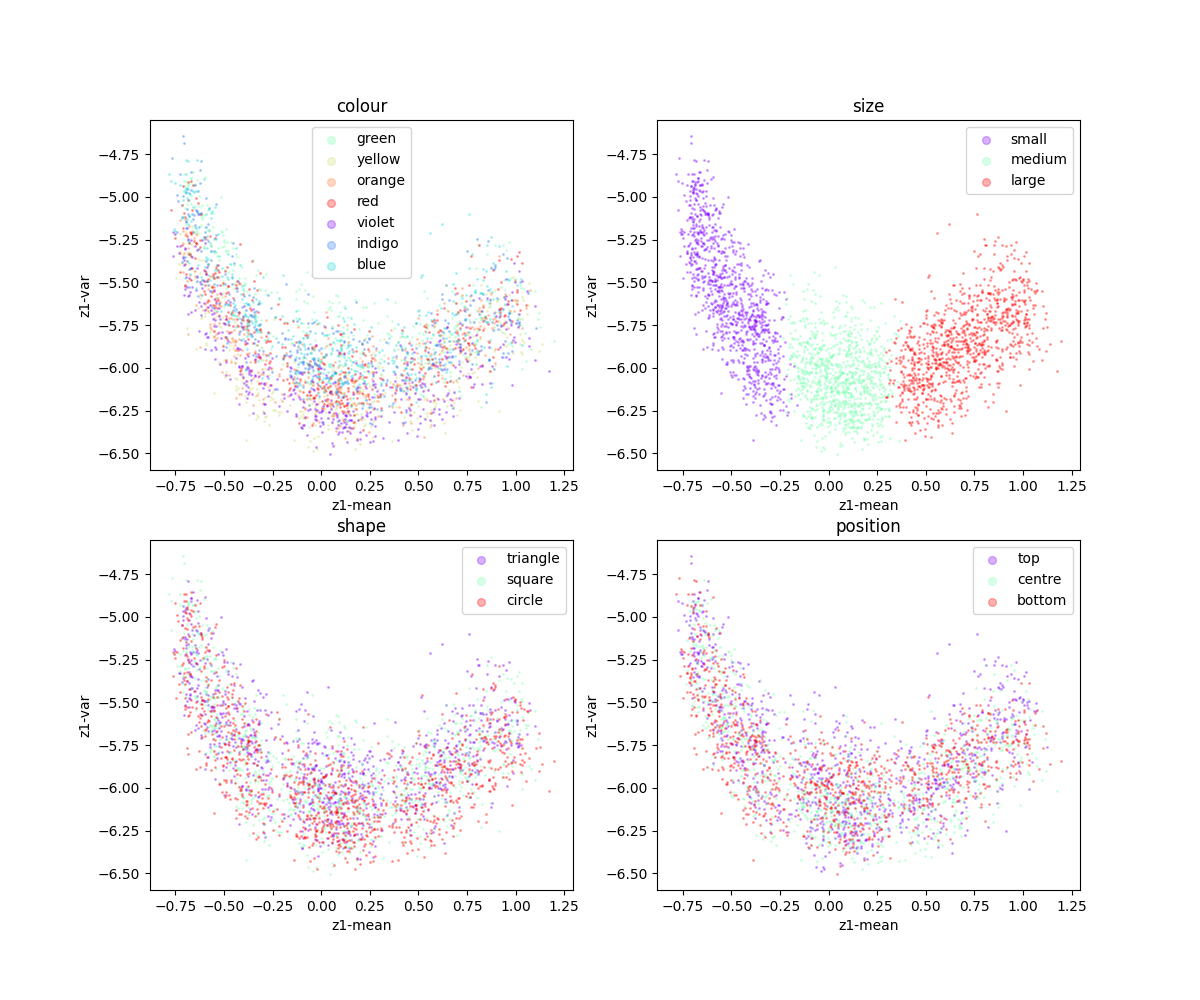}
    \hspace*{-0.5cm}\includegraphics[scale=0.22]{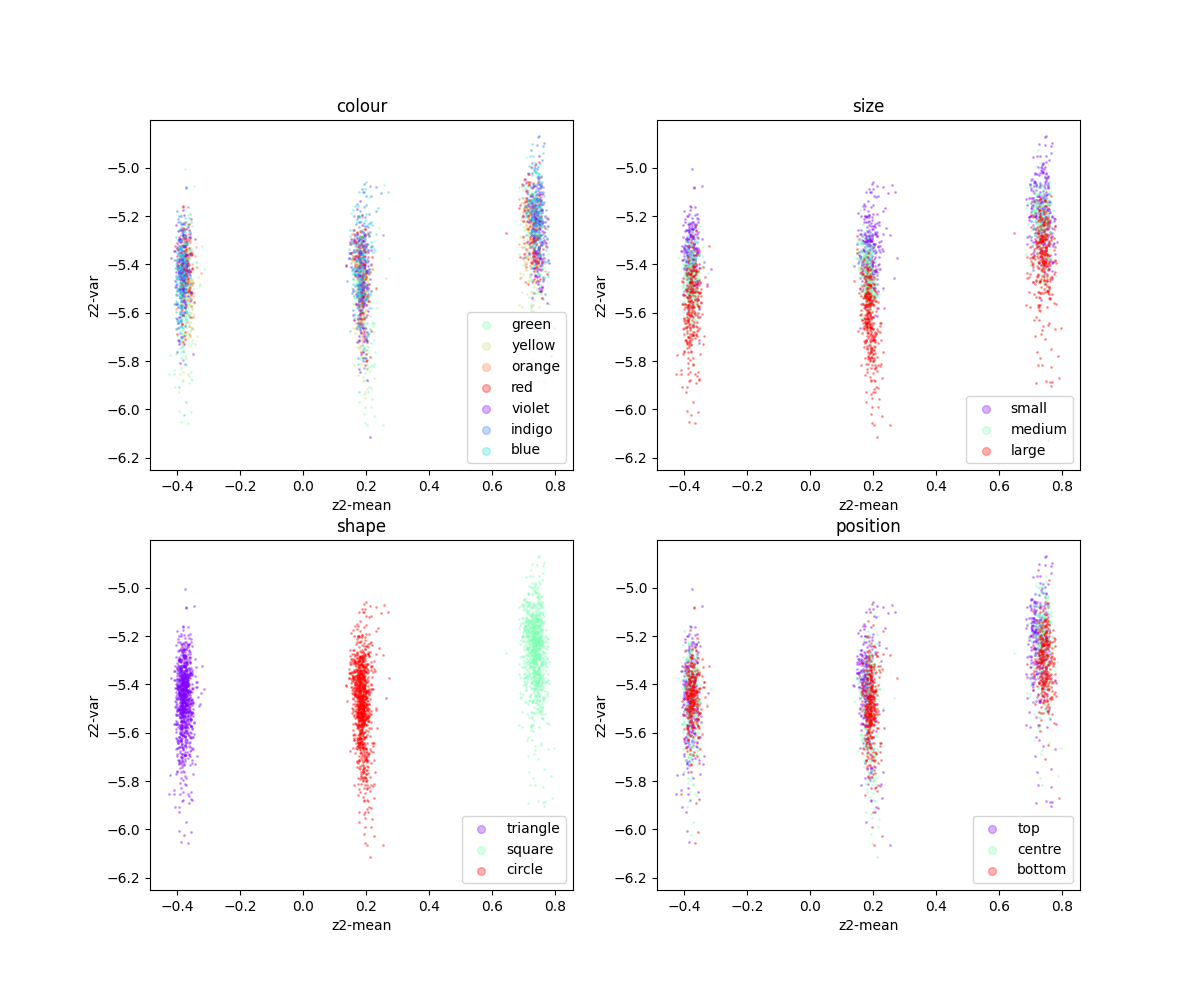}
    \includegraphics[scale=0.22]{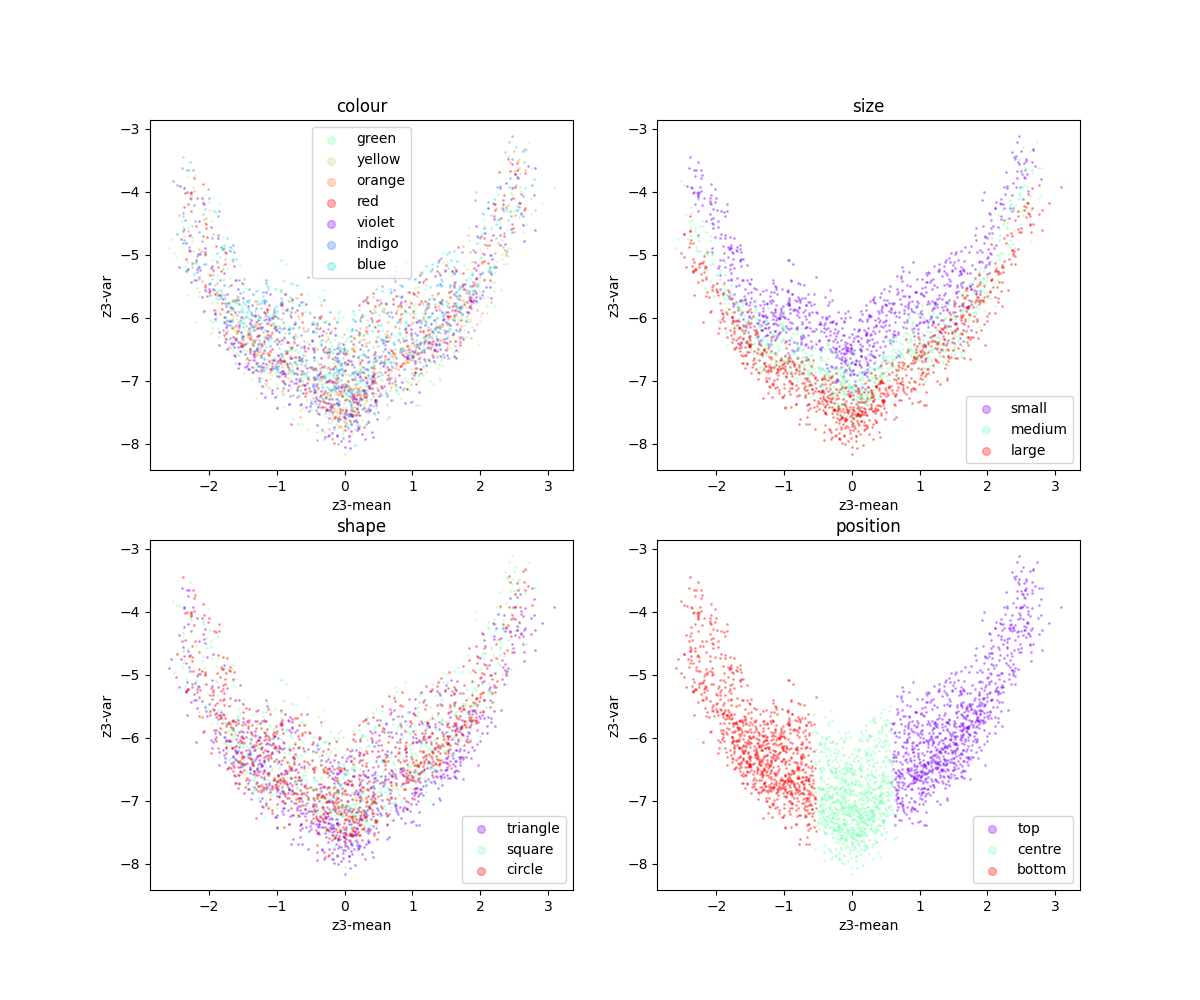}
    \caption{The means and log-variances for each dimension predicted by the encoder, for the rainbow extended-colour set.}
    \label{fig:clustering_rainbow}
\end{figure}

Again we chose a trained model which performed well on the development data (with accuracies well into the 90s for all domains), and plotted the colour-coded means and variances as predicted by the encoder.\footnote{No slack dimensions were used for this model.} Figure~\ref{fig:clustering_rainbow} again shows a neat separation for all the domains, with very similar patterns to those exhibited in Figure~\ref{fig:clustering}.
Appendix~\ref{sec:app_rainbow_clusters} contains an enlarged version of the plots from the top-left of Figure~\ref{fig:clustering_rainbow}. Looking carefully at the plot in the very top-left, we see that the colours are not only neatly separated along the \textsc{colour} dimension, but also that the ordering of the rainbow is faithfully represented: \emph{blue, indigo, violet, red, orange, yellow, green}. Anecdotally we have observed this behaviour consistently in a number of runs, but with different colours on the far left and far right of the plot.

\begin{figure}
    \centering
    \includegraphics[width=\textwidth]{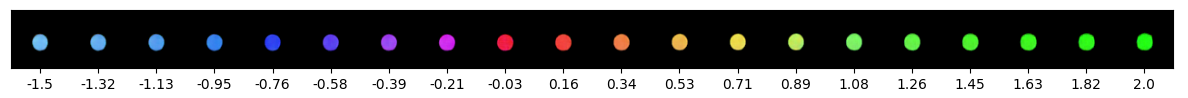}
    \includegraphics[width=\textwidth]{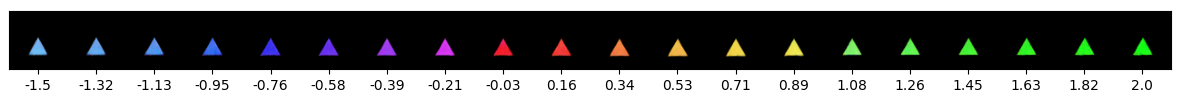}
    \caption{Traversals along the \textsc{colour} dimension for two examples from the colour-extended dataset.}
    \label{fig:traversal_example_rainbow}
\end{figure}

Figure~\ref{fig:traversal_example_rainbow} shows a couple of example traversals along the \textsc{colour} dimension only, for the colour-extended dataset, again demonstrating an ordering consistent with a rainbow.

\subsection{The \emph{any} Label}
\label{sec:any_expts}

Section~\ref{sec:fewer_labels} presented a model for learning from training instances in which only a subset of the conceptual domains are labelled with an atomic concept, with the remaining domains assigned the \emph{any} label. In order to test this model, we created a new training set of 6,000 instances (twice as many as previously), where each instance has 2 atomic labels and 2 \emph{any} labels, with the 2 labelled domains chosen at random. An example label is (\emph{any}, \emph{any}, \emph{circle}, \emph{top}). The 2 slack dimensions were retained, and 1,000 samples were used to give Monte Carlo estimates of the KLs for the Gaussian mixture models on the 2 dimensions labelled with \emph{any}. For the other 2 labelled dimensions the analytical expression for the KL was used, as before. 

Table~\ref{tab:classifier_results_any} (left) shows the accuracy of the classifier using this model, on the development and test data from Section~\ref{sec:classification}. (Again a model was chosen which performed well on the development data.) We do not show the cluster plots here, but the clusters for this model show the same patterns as in Figure~\ref{fig:clustering}, as would be expected for a model with such high classification accuracies.

\begin{table}[t]
    \centering
    \begin{tabular}{@{}l|rr|rr@{}}
    \toprule
    & \multicolumn{2}{c|}{2 \emph{any}'s}
    & \multicolumn{2}{c}{3 \emph{any}'s}\\
    & dev & test  & dev & test\\
    \hline
       \textsc{colour}  & 1.00 & 1.00 & 1.00 & 1.00 \\
     \textsc{size} &  0.99 & 0.98 & 0.97 & 0.95\\
     \textsc{shape} &  1.00 & 1.00 & 1.00 & 1.00\\
     \textsc{position} & 0.99 & 0.98 & 0.95 & 0.92\\
     \bottomrule
    \end{tabular}
    \caption{Classifier accuracy per domain for the Gaussian mixture model with the \emph{any} label.}
    \label{tab:classifier_results_any}
\end{table}

To test the model further, we created one more training set of 12,000 instances, this time with each instance assigned 3 \emph{any} labels (e.g. (\emph{any, any, any, centre})). This model was more difficult to train successfully, but we were able to train a model with the classification accuracies shown in Table~\ref{tab:classifier_results_any} (right). We introduced one more hyper-parameter which weights the parts of the KL loss corresponding to the \emph{any} label (with the dimensions labelled with atomic concepts weighted accordingly). For the model evaluated here, the part of the KL corresponding to the one atomic concept label was multiplied by 3.0, and each of the 3 dimensions labelled \emph{any} was multiplied by 0.333. The cluster plots still demonstrated a neat separation, but with slightly less of a coherent pattern for dimension 3 (\textsc{position}), which is to be expected given the lower test accuracy for that domain. Appendix~\ref{sec:app_dodgy_clusters} contains the plots.

\section{A Formal Model of Concepts}
\label{sec:formal}
Here we outline the theoretical background for our work, based on a formalisation of \Gardenfors' framework of \emph{conceptual spaces} \cite{gardenfors,gardenfors2014}, which models conceptual reasoning in human and artificial cognition. A central claim of \Gardenfors' framework is that concepts should be represented as convex regions of a space.
Conceptual space theory can be seen to incorporate aspects of the major psychological theories of concepts \cite{murphy_concepts}. In particular, the convex regions describing concepts can be formed from combining various instances or examples, as in \emph{exemplar theory} \cite{medin1978context}; they contain more ``central" or ``prototypical" points as in \emph{prototype theory} \cite{rosch1973natural}; and their geometric structure encodes underlying knowledge about the concepts, as in the \emph{knowledge theory} (or \emph{theory-theory}) \cite{murphy1985role}.

While various formalisations of the theory have been presented \cite{aisbett2001general,rickard2007reformulation,lewis2016hierarchical,bechberger2017thorough}, our work is motivated by the compositional approaches of \citeA{bolt2019interacting} and \citeA{tull2021categorical}, which we summarise here.\footnote{`Compositional' refers to the (many kinds of) composition in a monoidal category $(\catC, \otimes, \circ)$. Here we use only a specific form of composition - we view conceptual spaces as $\otimes$-products of domains, and consider concepts which factor over this product.}





\begin{definition} 
A \emph{conceptual space} is a set $Z$ in which we may take convex combinations of elements. That is, for all $z_1,\dots,z_n \in Z$ and $p_1, \dots, p_n \in [0,1]$ with $\sum^n_{i=1} p_i = 1$ there is an element of $Z$ denoted
\[
\sum^n_{i=1} p_i \cdot z_i 
\]
These convex combinations satisfy the rules one might expect; see \citeA{bolt2019interacting} for a more precise formulation.
Additionally we require that $Z$ forms a \emph{measure space}, meaning it comes with a $\sigma$-algebra $\Sigma_Z$ of measurable subsets $M \subseteq Z$, and a measure $\mu \colon \Sigma_Z \to [0,\infty]$.  
\end{definition} 

Typically a conceptual space is given as a product of simpler factors called \emph{domains}. That is, we have 
\[
Z = Z_1 \otimes \dots \otimes Z_n
\]
for domains $Z_1,\dots,Z_n$, which are themselves conceptual spaces.  Here $\otimes$ denotes the usual product $\times$ of sets (and measure spaces) with element-wise convex operations. 

Though they can be abstract\footnote{Note we do not use the bold font for $z$ in this section, since $z$ typically denotes an element of an abstract conceptual space and not necessarily a vector.}, all the conceptual spaces and domains we consider here are of the following concrete form.

\begin{example} \label{ex:RnCS}
Any convex subset $Z \subseteq \mathbb{R}^d$ forms a conceptual space. Here we equip $Z$ with the standard Lebesgue measure $\mu(A) = \int_A dz$ on $\mathbb{R}^d$. Thus any product $Z = Z_1 \times \dots \times Z_n$ of convex subsets $Z_i \subseteq \mathbb{R}^{d_i}$ forms a conceptual space also.
\end{example} 

Having defined conceptual spaces let us now consider concepts themselves. 

\begin{definition} 
A \emph{crisp concept} in a conceptual space $Z$ is a measurable  subset $\crc \subseteq Z$ which is \emph{convex}, meaning it is closed under convex combinations. 
\end{definition} 

We think of a point $z \in Z$ as belonging to the region $\crc$ of a crisp concept whenever it forms an instance of the concept. Convexity means that any point lying ``in-between" two instances will again belong to the concept. \Gardenfors{} justifies convexity based on cognitive experiments, including evidence from the division of colour space, and the relative ease of learning convex regions \cite{gardenfors}.

\begin{example} A simple example of a domain (from \citeA{bolt2019interacting}) is the \textsc{taste} simplex depicted below. This forms a convex subset of $\mathbb{R}^3$, generated by the extremal points \emph{sweet, bitter, salt} and \emph{sour}. Highlighted in red is a convex region describing a crisp concept for \emph{sweet}. 
\[
\includegraphics[scale=0.2]{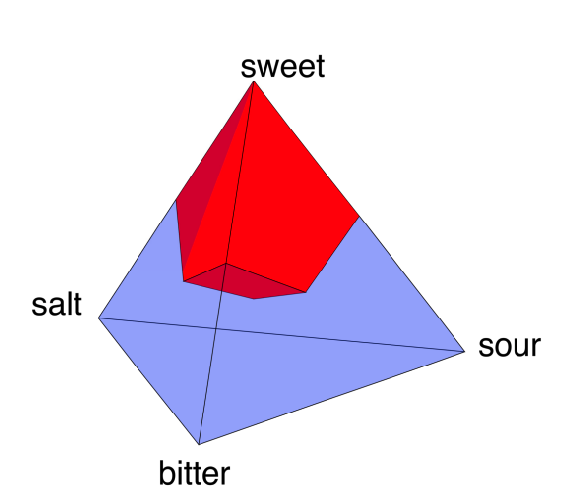}
\]
\end{example} 

\subsection{Fuzzy Concepts}

The concepts $\cc$ introduced so far have been \emph{crisp} in that each point $z$ strictly either satisfies $(z \in \cc)$ or fails to satisfy $(z \notin \cc)$ the concept. However, human concepts are typically understood to be \emph{fuzzy} or \emph{graded}, with membership taking some value in $[0,1]$. Fuzzy concepts are also convenient in the machine learning setting, such as in the VAEs explored here, readily allowing learning via gradient-based methods.

Formally, a fuzzy concept is given by a function 
\[
\cc \colon Z \to [0,1]
\]
For $z \in Z$, the value $\cc(z)$ represents the degree to which $z$ satisfies the concept, with $0$ meaning  not at all satisfied, and $1$ totally satisfied. If fuzzy concepts are to respect the structure of the conceptual space appropriately, they should not be arbitrary such mappings, but those which generalise convex subsets in some sense. \citeA{tull2021categorical} proposes the following definition:

\begin{definition}
A \emph{fuzzy concept} of $Z$ is a measurable function $\cc \colon Z \to [0,1]$ which is \emph{log-concave}, meaning that 
\[
\cc(p \cdot z + (1-p) \cdot z') \geq \cc(z)^p\cc(z')^{1-p}
\]
for all $z,z' \in Z$ and $p \in [0,1]$. 
\end{definition} 

In \citeA[Theorem 8]{tull2021categorical} it is proved that this is the most general definition of fuzzy concepts which satisfies a natural criterion due to \Gardenfors{}, includes crisp concepts (via their indicator functions $1_C$) and Gaussians, and is compositionally well-behaved in the following sense. 


\begin{lemma} \label{lem:product} 
Any product of fuzzy concepts $c_i \colon Z_i \to [0,1]$, for $i=1,\dots, n$ forms a fuzzy concept $c$ on $Z=Z_1 \times \dots Z_n$ via 
\begin{align*}
c (z_1,\dots,z_n) 
=c_1(z_1)\dots c_n(z_n) 
\end{align*} 
\end{lemma} 

Another common perspective on fuzzy concepts is to view them as probability distributions over $Z$. Formally, any fuzzy concept $\cc \colon Z \to [0,1]$ may, after suitable normalisation, be viewed as a density function for a probability measure
\begin{equation} \label{eq:fuzzy-density} 
p(z \mid \cc) := \frac{1}{\kappa} \cc(z) 
\end{equation} 
where $\kappa = \int_Z \cc(z) \,dz$ (provided this is non-zero and finite). The corresponding probability measure is then given by \[P(A\mid \cc) := \int_A p(z\mid \cc) \,dz\]
for each measurable subset $A \subseteq Z$. Intuitively, sampling from this distribution will produce points $z \in Z$ which are likely to fit the concept well, in that $p(z \mid \cc)$ (or equivalently $\cc(z)$) is high. 

In this work we consider only fuzzy concepts of the following form.

\begin{example} 
We may define a fuzzy concept $\cc \colon Z \to [0,1]$ on $Z=\mathbb{R}^n$ from any multivariate Gaussian 
\begin{align} \label{eq:Gaussian} 
\cc(z;\mu,\Sigma) &= e^{-\frac{1}{2}(z - \mu)^{\mathsf{T}}\Sigma^{-1}(z -\mu)} \\ 
& = e^{\sum^n_{i=1}-\frac{1}{2 \sigma_i^2}(z_i - \mu_i)^2}
\end{align}  
with mean $\mu$ and covariance matrix $\Sigma$. In the second line we restrict to the case where $\Sigma$ is diagonal, with $i$-th diagonal entry $\sigma_i^2$. Probabilistically this corresponds to a multivariate normal distribution $p(z \mid \cc)$ as in \eqref{eq:fuzzy-density} with $\kappa =\sqrt{\prod^n_{i=1} 2 \pi \sigma_i^2}$.

 Note that, in the case where $\Sigma$ is diagonal, any such Gaussian concept $\cc$ corresponds to a product of one-dimensional Gaussians, one per dimension, as in Lemma \ref{lem:product}:
 \[
 \cc(z; \mu, \Sigma) = \prod^n_{i=0} \cc_i(z_i;\mu_i,\sigma_i^2)
 \]
 These include the composite concepts in the conceptual VAE. For a concept such as $\cc=$(\emph{yellow, large, circle, top}), we have
\begin{equation} \label{eq:factored-concept-density}
p(z \mid \cc)
=
p(z_0 \mid \text{\emph {yellow}}) p(z_1 \mid \text{\emph{large}}) p(z_2 \mid \text{\emph{circle}}) p(z_3 \mid \text{\emph{top}})
\end{equation} 
where each $p(z_i \mid  \cc_i)$ is the density of a one-dimensional Gaussian.
 \end{example}

\section{Related Work}
\label{sec:related_work}

This work is inspired by \citeA{beta-vae}, who introduce the $\beta$-VAE for unsupervised concept learning. However, the focus of \citeauthor{beta-vae} is on learning the conceptual \emph{domains}, i.e. the underlying factors generating the data \shortcite{bengio:2013}, which they refer to as learning a \emph{disentangled} representation. The main innovation to encourage the VAE to learn a disentangled, or factored, latent space is the introduction of a weighting term $\beta$ on the KL loss. \citeauthor{beta-vae} show that setting $\beta$ to a value greater than 1 can result in the dimensions of $\Z$ corresponding to domains such as the lighting or elevation of a face in the celebA images dataset, or the width of a chair in a dataset of chair images.

Our focus is more on the conceptual representations themselves, assuming the domains are already known, and the question of how concept labels can be introduced into the VAE model. \citeA{SCAN} also consider how labels can be associated with concepts, but again with a focus on the unsupervised learning of the underlying factors of the latent space together with the conceptual primitives which make up a conceptual hierarchy (such as individual colours and sizes). \citeA{clark_concepts}, as part of a theoretical study, also suggest how a conceptual hierarchy could be learned from the output of a VAE encoder $q(\z|\x)$ when applied to a set of images $\X$.

A recent trend in the NLP word embedding literature, building on the original work of \shortciteA{word2vec} and \shortciteA{glove}, is to consider probabilistic word embeddings represented as densities, including Gaussians \cite{vilnis}. \citeA{nickel} consider the geometry of the embedding space, and argue for a hyperbolic, rather than Euclidean, geometry. A paper in NLP that uses a model very similar to ours is \shortciteA{brazinskas} which introduces the Bayesian skip-gram model for learning word embeddings. One key difference which distinguishes our work from the word embeddings typically used in NLP is that we do not restrict ourselves to the textual domain, meaning that our conceptual representations are \emph{grounded} in some other modality (in our case images) \cite{harnad}, bringing them closer to the human conceptual system.


Here we have used Gaussians to represent concepts, since they are the typical distributions used with VAEs and because they are convenient from a mathematical perspective. However, the use of Gaussians is also prevalent in the neuroscience literature, appearing for example  as the \emph{Laplace assumption} in the ``free-energy" or ``predictive processing" framework \cite{friston2009predictive,bogacz2017tutorial}.   
\section{Conclusion and Further Work}
\label{sec:further_work}

In this report we have presented a new model of concepts based on the VAE framework, showing how it can be trained on images to induce interpretable concepts which can be used for e.g. concept classification. The obvious extension to this work is to apply it to more realistic images, with more domains and concepts. However, the level of supervision that we have provided, in terms of the conceptual domains, is unlikely to scale. Hence one promising future direction is to combine the Conceptual VAE presented here with the unsupervised learning of domains from the $\beta$-VAE (or more recent models such as \citeA{leeb}). Learning concept labels and domains from large bodies of text is another promising avenue for future work, building on the extensive NLP literature for inducing conceptual hierarchies (e.g. \citeA{pasca-van-durme-2008-weakly}).

There are many quantitative aspects of the model left to explore. For example, we have not investigated the question of sample efficiency, and whether the model can still effectively learn when provided with (much) less data. Similarly, it would be interesting to investigate the generalisation capabilities of the model, and in particular whether it can classify compound concepts ``zero-shot", based on experience of the component atomic concepts (e.g. can the model recognize a red circle when it has only seen red squares and blue circles). The structure inherent in the Conceptual VAE suggests that it ought to perform well on such out-of-distribution test cases.

Another interesting question is whether the Conceptual VAE can be applied to data generated from a conceptual hierarchy---for example having shades of colour such as \emph{dark-red}---and whether the learned Gaussian representations for concepts can be partially ordered in an appropriate way \cite{clark_concepts}. And continuing with the representation of colour, here we have only modelled the hue---along a single dimension---whereas colour is more appropriately represented using something like a colour spindle. Hence it is likely that \textsc{colour} (along with many other domains) will require more than one latent dimension, with some appropriate structure, to be modelled correctly.

The concepts considered here are all given as a product of one concept over each domain, as in \eqref{eq:factored-concept-density}. In future we hope to explore the learning of non-factored concepts, which may relate domains. For example, we might imagine a \emph{rainbow} concept, which correlates \textsc{position} and \textsc{colour} according to the colours of the rainbow. A more realistic example would be the concept of \emph{banana} for which \textsc{sweetness} and \textsc{colour} are correlated.



In \citeA{tull2021categorical} the definition of fuzzy concepts is extended to define fuzzy \emph{conceptual processes} between conceptual spaces. It would be interesting to explore the learning of such processes, including ``metaphorical" mappings between domains. Fuzzy conceptual processes form a \emph{symmetric monoidal category}, allowing one to reason about and compose concepts using simple \emph{string diagrams}. These were previously explored for crisp concepts in \citeA{bolt2019interacting} to give conceptual semantics for natural language, by converting sentences into concepts diagrammatically (see (\ref{eq:DisCoDiag}) below). In future we hope to explore the effectiveness of similar diagrammatic composition procedures for learned fuzzy concepts. 
\begin{equation} \label{eq:DisCoDiag}
\includegraphics[scale=0.18]{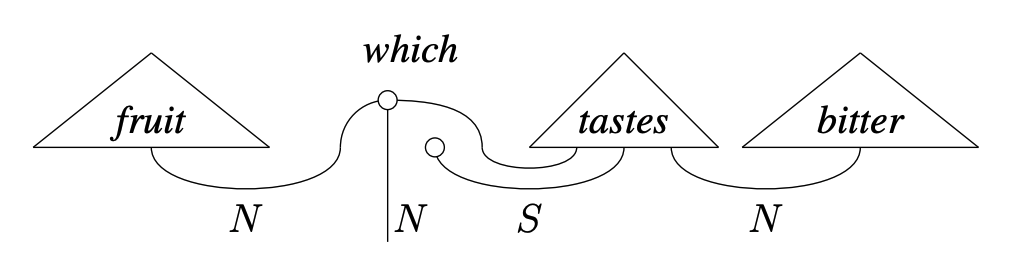}
= 
\includegraphics[scale=0.18]{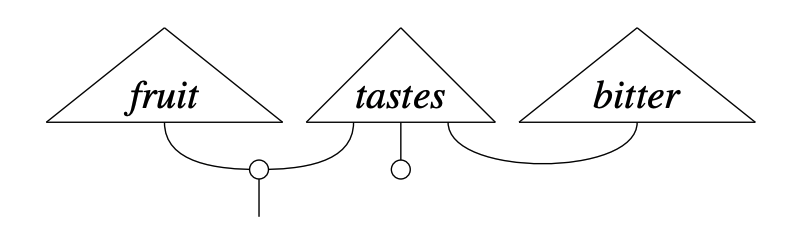}
\end{equation} 

In addition, this diagrammatic formalism applies especially well to quantum theory and quantum computation \cite{dodo_book}. It would be interesting to extend our setup to a quantum model, perhaps using a variant of the quantum VAE \cite{khoshaman2018quantum}, and incorporating techniques from the literature on VAEs with discrete latent representations \shortcite{discrete-vae}.

\section*{Acknowledgements}

SC would like to thank the DeepMind Concepts Team, with whom he worked for a year before joining CQ, and from whom he took inspiration for many of the ideas in this report; and also Chris Dyer from the DeepMind Language team, who gave SC the idea of introducing a word label at the top of the VAE. We would also like to thank Bob Coecke, Vincent Wang-Ma\'scianica, Robin Lorenz, Konstantinos Meichanetzidis and the whole of the Quantum Compositional Intelligence team at CQ for useful feedback on this work.

\bibliographystyle{apacite}

\bibliography{references}

\begin{thebibliography}{}

\bibitem [\protect \citeauthoryear {%
Aisbett%
\ \BBA {} Gibbon%
}{%
Aisbett%
\ \BBA {} Gibbon%
}{%
{\protect \APACyear {2001}}%
}]{%
aisbett2001general}
\APACinsertmetastar {%
aisbett2001general}%
\begin{APACrefauthors}%
Aisbett, J.%
\BCBT {}\ \BBA {} Gibbon, G.%
\end{APACrefauthors}%
\unskip\
\newblock
\APACrefYearMonthDay{2001}{}{}.
\newblock
{\BBOQ}\APACrefatitle {A general formulation of conceptual spaces as a meso
  level representation} {A general formulation of conceptual spaces as a meso
  level representation}.{\BBCQ}
\newblock
\APACjournalVolNumPages{Artificial Intelligence}{133}{1-2}{189--232}.
\PrintBackRefs{\CurrentBib}

\bibitem [\protect \citeauthoryear {%
Bechberger%
\ \BBA {} K{\"u}hnberger%
}{%
Bechberger%
\ \BBA {} K{\"u}hnberger%
}{%
{\protect \APACyear {2017}}%
}]{%
bechberger2017thorough}
\APACinsertmetastar {%
bechberger2017thorough}%
\begin{APACrefauthors}%
Bechberger, L.%
\BCBT {}\ \BBA {} K{\"u}hnberger, K\BHBI U.%
\end{APACrefauthors}%
\unskip\
\newblock
\APACrefYearMonthDay{2017}{}{}.
\newblock
{\BBOQ}\APACrefatitle {A thorough formalization of conceptual spaces} {A
  thorough formalization of conceptual spaces}.{\BBCQ}
\newblock
\BIn{} \APACrefbtitle {Joint German/Austrian Conference on Artificial
  Intelligence (K{\"u}nstliche Intelligenz)} {Joint german/austrian conference
  on artificial intelligence (k{\"u}nstliche intelligenz)}\ (\BPGS\ 58--71).
\PrintBackRefs{\CurrentBib}

\bibitem [\protect \citeauthoryear {%
Bengio%
, Courville%
\BCBL {}\ \BBA {} Vincent%
}{%
Bengio%
\ \protect \BOthers {.}}{%
{\protect \APACyear {2013}}%
}]{%
bengio:2013}
\APACinsertmetastar {%
bengio:2013}%
\begin{APACrefauthors}%
Bengio, Y.%
, Courville, A.%
\BCBL {}\ \BBA {} Vincent, P.%
\end{APACrefauthors}%
\unskip\
\newblock
\APACrefYearMonthDay{2013}{}{}.
\newblock
{\BBOQ}\APACrefatitle {Representation learning: A review and new perspectives}
  {Representation learning: A review and new perspectives}.{\BBCQ}
\newblock
\APACjournalVolNumPages{IEEE Transactions on Pattern Analysis \& Machine
  Intelligence}{}{}{}.
\PrintBackRefs{\CurrentBib}

\bibitem [\protect \citeauthoryear {%
Bogacz%
}{%
Bogacz%
}{%
{\protect \APACyear {2017}}%
}]{%
bogacz2017tutorial}
\APACinsertmetastar {%
bogacz2017tutorial}%
\begin{APACrefauthors}%
Bogacz, R.%
\end{APACrefauthors}%
\unskip\
\newblock
\APACrefYearMonthDay{2017}{}{}.
\newblock
{\BBOQ}\APACrefatitle {A tutorial on the free-energy framework for modelling
  perception and learning} {A tutorial on the free-energy framework for
  modelling perception and learning}.{\BBCQ}
\newblock
\APACjournalVolNumPages{Journal of mathematical psychology}{76}{}{198--211}.
\PrintBackRefs{\CurrentBib}

\bibitem [\protect \citeauthoryear {%
Bolt%
\ \protect \BOthers {.}}{%
Bolt%
\ \protect \BOthers {.}}{%
{\protect \APACyear {2019}}%
}]{%
bolt2019interacting}
\APACinsertmetastar {%
bolt2019interacting}%
\begin{APACrefauthors}%
Bolt, J.%
, Coecke, B.%
, Genovese, F.%
, Lewis, M.%
, Marsden, D.%
\BCBL {}\ \BBA {} Piedeleu, R.%
\end{APACrefauthors}%
\unskip\
\newblock
\APACrefYearMonthDay{2019}{}{}.
\newblock
{\BBOQ}\APACrefatitle {Interacting conceptual spaces I: Grammatical composition
  of concepts} {Interacting conceptual spaces i: Grammatical composition of
  concepts}.{\BBCQ}
\newblock
\BIn{} \APACrefbtitle {Conceptual Spaces: Elaborations and Applications}
  {Conceptual spaces: Elaborations and applications}\ (\BPGS\ 151--181).
\newblock
\APACaddressPublisher{}{Springer}.
\PrintBackRefs{\CurrentBib}

\bibitem [\protect \citeauthoryear {%
Bra{\v{z}}inskas%
, Havrylov%
\BCBL {}\ \BBA {} Titov%
}{%
Bra{\v{z}}inskas%
\ \protect \BOthers {.}}{%
{\protect \APACyear {2018}}%
}]{%
brazinskas}
\APACinsertmetastar {%
brazinskas}%
\begin{APACrefauthors}%
Bra{\v{z}}inskas, A.%
, Havrylov, S.%
\BCBL {}\ \BBA {} Titov, I.%
\end{APACrefauthors}%
\unskip\
\newblock
\APACrefYearMonthDay{2018}{{\APACmonth{08}}}{}.
\newblock
{\BBOQ}\APACrefatitle {Embedding Words as Distributions with a {B}ayesian
  Skip-gram Model} {Embedding words as distributions with a {B}ayesian
  skip-gram model}.{\BBCQ}
\newblock
\BIn{} \APACrefbtitle {Proceedings of the 27th International Conference on
  Computational Linguistics} {Proceedings of the 27th international conference
  on computational linguistics}\ (\BPGS\ 1775--1789).
\newblock
\APACaddressPublisher{Santa Fe, New Mexico, USA}{Association for Computational
  Linguistics}.
\newblock
\begin{APACrefURL} \url{https://aclanthology.org/C18-1151} \end{APACrefURL}
\PrintBackRefs{\CurrentBib}

\bibitem [\protect \citeauthoryear {%
Clark%
\ \protect \BOthers {.}}{%
Clark%
\ \protect \BOthers {.}}{%
{\protect \APACyear {2021}}%
}]{%
clark_concepts}
\APACinsertmetastar {%
clark_concepts}%
\begin{APACrefauthors}%
Clark, S.%
, Lerchner, A.%
, von Glehn, T.%
, Tieleman, O.%
, Tanburn, R.%
, Dashevskiy, M.%
\BCBL {}\ \BBA {} Bosnjak, M.%
\end{APACrefauthors}%
\unskip\
\newblock
\APACrefYearMonthDay{2021}{}{}.
\newblock
\APACrefbtitle {Formalising Concepts as Grounded Abstractions} {Formalising
  concepts as grounded abstractions}\ \APACbVolEdTR{}{\BTR{}}.
\newblock
\APACaddressInstitution{https://arxiv.org/pdf/2101.05125.pdf}{DeepMind,
  London}.
\PrintBackRefs{\CurrentBib}

\bibitem [\protect \citeauthoryear {%
Coecke%
\ \BBA {} Kissinger%
}{%
Coecke%
\ \BBA {} Kissinger%
}{%
{\protect \APACyear {2017}}%
}]{%
dodo_book}
\APACinsertmetastar {%
dodo_book}%
\begin{APACrefauthors}%
Coecke, B.%
\BCBT {}\ \BBA {} Kissinger, A.%
\end{APACrefauthors}%
\unskip\
\newblock
\APACrefYear{2017}.
\newblock
\APACrefbtitle {Picturing Quantum Processes - A First Course in Quantum Theory
  and Diagrammatic Reasoning} {Picturing quantum processes - a first course in
  quantum theory and diagrammatic reasoning}.
\newblock
\APACaddressPublisher{}{Cambridge University Press}.
\PrintBackRefs{\CurrentBib}

\bibitem [\protect \citeauthoryear {%
Doersch%
}{%
Doersch%
}{%
{\protect \APACyear {2016}}%
}]{%
doersch}
\APACinsertmetastar {%
doersch}%
\begin{APACrefauthors}%
Doersch, C.%
\end{APACrefauthors}%
\unskip\
\newblock
\APACrefYearMonthDay{2016}{}{}.
\newblock
\APACrefbtitle {Tutorial on Variational Autoencoders} {Tutorial on variational
  autoencoders}\ \APACbVolEdTR{}{\BTR{}}.
\newblock
\APACaddressInstitution{https://arxiv.org/abs/1606.05908}{}.
\PrintBackRefs{\CurrentBib}

\bibitem [\protect \citeauthoryear {%
Evans%
}{%
Evans%
}{%
{\protect \APACyear {2019}}%
}]{%
evans}
\APACinsertmetastar {%
evans}%
\begin{APACrefauthors}%
Evans, V.%
\end{APACrefauthors}%
\unskip\
\newblock
\APACrefYear{2019}.
\newblock
\APACrefbtitle {Cognitive Linguistics - A Complete Guide (Second Edition)}
  {Cognitive linguistics - a complete guide (second edition)}.
\newblock
\APACaddressPublisher{}{Edinburgh University Press}.
\PrintBackRefs{\CurrentBib}

\bibitem [\protect \citeauthoryear {%
Friston%
\ \BBA {} Kiebel%
}{%
Friston%
\ \BBA {} Kiebel%
}{%
{\protect \APACyear {2009}}%
}]{%
friston2009predictive}
\APACinsertmetastar {%
friston2009predictive}%
\begin{APACrefauthors}%
Friston, K.%
\BCBT {}\ \BBA {} Kiebel, S.%
\end{APACrefauthors}%
\unskip\
\newblock
\APACrefYearMonthDay{2009}{}{}.
\newblock
{\BBOQ}\APACrefatitle {Predictive coding under the free-energy principle}
  {Predictive coding under the free-energy principle}.{\BBCQ}
\newblock
\APACjournalVolNumPages{Philosophical transactions of the Royal Society B:
  Biological sciences}{364}{1521}{1211--1221}.
\PrintBackRefs{\CurrentBib}

\bibitem [\protect \citeauthoryear {%
Ganter%
\ \BBA {} Obiedkov%
}{%
Ganter%
\ \BBA {} Obiedkov%
}{%
{\protect \APACyear {2016}}%
}]{%
Ganter2016}
\APACinsertmetastar {%
Ganter2016}%
\begin{APACrefauthors}%
Ganter, B.%
\BCBT {}\ \BBA {} Obiedkov, S.%
\end{APACrefauthors}%
\unskip\
\newblock
\APACrefYear{2016}.
\newblock
\APACrefbtitle {Conceptual Exploration} {Conceptual exploration}.
\newblock
\APACaddressPublisher{}{Springer}.
\PrintBackRefs{\CurrentBib}

\bibitem [\protect \citeauthoryear {%
G{\"a}rdenfors%
}{%
G{\"a}rdenfors%
}{%
{\protect \APACyear {2000}}%
}]{%
gardenfors}
\APACinsertmetastar {%
gardenfors}%
\begin{APACrefauthors}%
G{\"a}rdenfors, P.%
\end{APACrefauthors}%
\unskip\
\newblock
\APACrefYear{2000}.
\newblock
\APACrefbtitle {Conceptual Spaces: The Geometry of Thought} {Conceptual spaces:
  The geometry of thought}.
\newblock
\APACaddressPublisher{}{The MIT Press}.
\PrintBackRefs{\CurrentBib}

\bibitem [\protect \citeauthoryear {%
G{\"a}rdenfors%
}{%
G{\"a}rdenfors%
}{%
{\protect \APACyear {2014}}%
}]{%
gardenfors2014}
\APACinsertmetastar {%
gardenfors2014}%
\begin{APACrefauthors}%
G{\"a}rdenfors, P.%
\end{APACrefauthors}%
\unskip\
\newblock
\APACrefYear{2014}.
\newblock
\APACrefbtitle {The Geometry of Meaning} {The geometry of meaning}.
\newblock
\APACaddressPublisher{}{The MIT Press}.
\PrintBackRefs{\CurrentBib}

\bibitem [\protect \citeauthoryear {%
Harnad%
}{%
Harnad%
}{%
{\protect \APACyear {1990}}%
}]{%
harnad}
\APACinsertmetastar {%
harnad}%
\begin{APACrefauthors}%
Harnad, S.%
\end{APACrefauthors}%
\unskip\
\newblock
\APACrefYearMonthDay{1990}{}{}.
\newblock
{\BBOQ}\APACrefatitle {The symbol grounding problem} {The symbol grounding
  problem}.{\BBCQ}
\newblock
\APACjournalVolNumPages{Physica D: Nonlinear Phenomona}{42}{}{335-346}.
\PrintBackRefs{\CurrentBib}

\bibitem [\protect \citeauthoryear {%
Hearst%
}{%
Hearst%
}{%
{\protect \APACyear {1992}}%
}]{%
hearst}
\APACinsertmetastar {%
hearst}%
\begin{APACrefauthors}%
Hearst, M\BPBI A.%
\end{APACrefauthors}%
\unskip\
\newblock
\APACrefYearMonthDay{1992}{}{}.
\newblock
{\BBOQ}\APACrefatitle {Automatic acquisition of hyponyms from large text
  corpora} {Automatic acquisition of hyponyms from large text corpora}.{\BBCQ}
\newblock
\BIn{} \APACrefbtitle {Proceedings of the 1992 Conference on Computational
  Linguistics.} {Proceedings of the 1992 conference on computational
  linguistics.}
\PrintBackRefs{\CurrentBib}

\bibitem [\protect \citeauthoryear {%
Higgins%
\ \protect \BOthers {.}}{%
Higgins%
\ \protect \BOthers {.}}{%
{\protect \APACyear {2017}}%
}]{%
beta-vae}
\APACinsertmetastar {%
beta-vae}%
\begin{APACrefauthors}%
Higgins, I.%
, Matthey, L.%
, Pal, A.%
, Burgess, C\BPBI P.%
, Glorot, X.%
, Botvinick, M.%
\BDBL {}Lerchner, A.%
\end{APACrefauthors}%
\unskip\
\newblock
\APACrefYearMonthDay{2017}{}{}.
\newblock
{\BBOQ}\APACrefatitle {$\beta$-{VAE}: Learning Basic Visual Concepts with a
  Constrained Variational Framework} {$\beta$-{VAE}: Learning basic visual
  concepts with a constrained variational framework}.{\BBCQ}
\newblock
\BIn{} \APACrefbtitle {Proceedings of {ICLR} 2017.} {Proceedings of {ICLR}
  2017.}
\PrintBackRefs{\CurrentBib}

\bibitem [\protect \citeauthoryear {%
Higgins%
\ \protect \BOthers {.}}{%
Higgins%
\ \protect \BOthers {.}}{%
{\protect \APACyear {2018}}%
}]{%
SCAN}
\APACinsertmetastar {%
SCAN}%
\begin{APACrefauthors}%
Higgins, I.%
, Sonnerat, N.%
, Matthey, L.%
, Pal, A.%
, Burgess, C\BPBI P.%
, Bo\v{s}njak, M.%
\BDBL {}Lerchner, A.%
\end{APACrefauthors}%
\unskip\
\newblock
\APACrefYearMonthDay{2018}{}{}.
\newblock
{\BBOQ}\APACrefatitle {{SCAN}: Learning Hierarchical Compositional Visual
  Concepts} {{SCAN}: Learning hierarchical compositional visual
  concepts}.{\BBCQ}
\newblock
\BIn{} \APACrefbtitle {Proceedings of {ICLR} 2018.} {Proceedings of {ICLR}
  2018.}
\PrintBackRefs{\CurrentBib}

\bibitem [\protect \citeauthoryear {%
Khoshaman%
\ \protect \BOthers {.}}{%
Khoshaman%
\ \protect \BOthers {.}}{%
{\protect \APACyear {2018}}%
}]{%
khoshaman2018quantum}
\APACinsertmetastar {%
khoshaman2018quantum}%
\begin{APACrefauthors}%
Khoshaman, A.%
, Vinci, W.%
, Denis, B.%
, Andriyash, E.%
, Sadeghi, H.%
\BCBL {}\ \BBA {} Amin, M\BPBI H.%
\end{APACrefauthors}%
\unskip\
\newblock
\APACrefYearMonthDay{2018}{}{}.
\newblock
{\BBOQ}\APACrefatitle {Quantum variational autoencoder} {Quantum variational
  autoencoder}.{\BBCQ}
\newblock
\APACjournalVolNumPages{Quantum Science and Technology}{4}{1}{014001}.
\PrintBackRefs{\CurrentBib}

\bibitem [\protect \citeauthoryear {%
Kingma%
\ \BBA {} Welling%
}{%
Kingma%
\ \BBA {} Welling%
}{%
{\protect \APACyear {2014}}%
}]{%
kingma14}
\APACinsertmetastar {%
kingma14}%
\begin{APACrefauthors}%
Kingma, D\BPBI P.%
\BCBT {}\ \BBA {} Welling, M.%
\end{APACrefauthors}%
\unskip\
\newblock
\APACrefYearMonthDay{2014}{}{}.
\newblock
{\BBOQ}\APACrefatitle {Auto-Encoding Variational {B}ayes} {Auto-encoding
  variational {B}ayes}.{\BBCQ}
\newblock
\BIn{} \APACrefbtitle {Proceedings of the International Conference on Learning
  Representations ({ICLR} 2014).} {Proceedings of the international conference
  on learning representations ({ICLR} 2014).}
\PrintBackRefs{\CurrentBib}

\bibitem [\protect \citeauthoryear {%
Lake%
, Ullman%
, Tenenbaum%
\BCBL {}\ \BBA {} Gershman%
}{%
Lake%
\ \protect \BOthers {.}}{%
{\protect \APACyear {2017}}%
}]{%
lake_thinking_machines}
\APACinsertmetastar {%
lake_thinking_machines}%
\begin{APACrefauthors}%
Lake, B\BPBI M.%
, Ullman, T\BPBI D.%
, Tenenbaum, J\BPBI B.%
\BCBL {}\ \BBA {} Gershman, S\BPBI J.%
\end{APACrefauthors}%
\unskip\
\newblock
\APACrefYearMonthDay{2017}{}{}.
\newblock
{\BBOQ}\APACrefatitle {Building machines that learn and think like people}
  {Building machines that learn and think like people}.{\BBCQ}
\newblock
\APACjournalVolNumPages{Behavioral and Brain Sciences}{40}{}{}.
\PrintBackRefs{\CurrentBib}

\bibitem [\protect \citeauthoryear {%
Leeb%
\ \protect \BOthers {.}}{%
Leeb%
\ \protect \BOthers {.}}{%
{\protect \APACyear {2021}}%
}]{%
leeb}
\APACinsertmetastar {%
leeb}%
\begin{APACrefauthors}%
Leeb, F.%
, Lanzillotta, G.%
, Annadani, Y.%
, Besserve, M.%
, Bauer, S.%
\BCBL {}\ \BBA {} Sch{\"o}lkopf, B.%
\end{APACrefauthors}%
\unskip\
\newblock
\APACrefYearMonthDay{2021}{}{}.
\newblock
\APACrefbtitle {Structure by Architecture: Disentangled Representations without
  Regularization.} {Structure by architecture: Disentangled representations
  without regularization.}
\newblock
\APAChowpublished {https://arxiv.org/abs/2006.07796}.
\PrintBackRefs{\CurrentBib}

\bibitem [\protect \citeauthoryear {%
Lewis%
\ \BBA {} Lawry%
}{%
Lewis%
\ \BBA {} Lawry%
}{%
{\protect \APACyear {2016}}%
}]{%
lewis2016hierarchical}
\APACinsertmetastar {%
lewis2016hierarchical}%
\begin{APACrefauthors}%
Lewis, M.%
\BCBT {}\ \BBA {} Lawry, J.%
\end{APACrefauthors}%
\unskip\
\newblock
\APACrefYearMonthDay{2016}{}{}.
\newblock
{\BBOQ}\APACrefatitle {Hierarchical conceptual spaces for concept combination}
  {Hierarchical conceptual spaces for concept combination}.{\BBCQ}
\newblock
\APACjournalVolNumPages{Artificial Intelligence}{237}{}{204--227}.
\PrintBackRefs{\CurrentBib}

\bibitem [\protect \citeauthoryear {%
Margolis%
\ \BBA {} Laurence%
}{%
Margolis%
\ \BBA {} Laurence%
}{%
{\protect \APACyear {2015}}%
}]{%
conceptual_mind}
\APACinsertmetastar {%
conceptual_mind}%
\begin{APACrefauthors}%
Margolis, E.%
\BCBT {}\ \BBA {} Laurence, S.%
\end{APACrefauthors}%
\ (\BEDS).
\unskip\
\newblock
\APACrefYear{2015}.
\newblock
\APACrefbtitle {The Conceptual Mind: New Directions in the Study of Concepts}
  {The conceptual mind: New directions in the study of concepts}.
\newblock
\APACaddressPublisher{}{The MIT Press}.
\PrintBackRefs{\CurrentBib}

\bibitem [\protect \citeauthoryear {%
Medin%
\ \BBA {} Schaffer%
}{%
Medin%
\ \BBA {} Schaffer%
}{%
{\protect \APACyear {1978}}%
}]{%
medin1978context}
\APACinsertmetastar {%
medin1978context}%
\begin{APACrefauthors}%
Medin, D\BPBI L.%
\BCBT {}\ \BBA {} Schaffer, M\BPBI M.%
\end{APACrefauthors}%
\unskip\
\newblock
\APACrefYearMonthDay{1978}{}{}.
\newblock
{\BBOQ}\APACrefatitle {Context theory of classification learning.} {Context
  theory of classification learning.}{\BBCQ}
\newblock
\APACjournalVolNumPages{Psychological review}{85}{3}{207}.
\PrintBackRefs{\CurrentBib}

\bibitem [\protect \citeauthoryear {%
Mikolov%
, Chen%
, Corrado%
\BCBL {}\ \BBA {} Dean%
}{%
Mikolov%
\ \protect \BOthers {.}}{%
{\protect \APACyear {2013}}%
}]{%
word2vec}
\APACinsertmetastar {%
word2vec}%
\begin{APACrefauthors}%
Mikolov, T.%
, Chen, K.%
, Corrado, G.%
\BCBL {}\ \BBA {} Dean, J.%
\end{APACrefauthors}%
\unskip\
\newblock
\APACrefYearMonthDay{2013}{}{}.
\newblock
{\BBOQ}\APACrefatitle {Efficient estimation of word representations in vector
  space} {Efficient estimation of word representations in vector space}.{\BBCQ}
\newblock
\BIn{} \APACrefbtitle {Proceedings of {ICLR} 2013 Workshop.} {Proceedings of
  {ICLR} 2013 workshop.}
\PrintBackRefs{\CurrentBib}

\bibitem [\protect \citeauthoryear {%
Murphy%
}{%
Murphy%
}{%
{\protect \APACyear {2002}}%
}]{%
murphy_concepts}
\APACinsertmetastar {%
murphy_concepts}%
\begin{APACrefauthors}%
Murphy, G\BPBI L.%
\end{APACrefauthors}%
\unskip\
\newblock
\APACrefYear{2002}.
\newblock
\APACrefbtitle {The Big Book of Concepts} {The big book of concepts}.
\newblock
\APACaddressPublisher{}{The MIT Press}.
\PrintBackRefs{\CurrentBib}

\bibitem [\protect \citeauthoryear {%
Murphy%
\ \BBA {} Medin%
}{%
Murphy%
\ \BBA {} Medin%
}{%
{\protect \APACyear {1985}}%
}]{%
murphy1985role}
\APACinsertmetastar {%
murphy1985role}%
\begin{APACrefauthors}%
Murphy, G\BPBI L.%
\BCBT {}\ \BBA {} Medin, D\BPBI L.%
\end{APACrefauthors}%
\unskip\
\newblock
\APACrefYearMonthDay{1985}{}{}.
\newblock
{\BBOQ}\APACrefatitle {The role of theories in conceptual coherence.} {The role
  of theories in conceptual coherence.}{\BBCQ}
\newblock
\APACjournalVolNumPages{Psychological review}{92}{3}{289}.
\PrintBackRefs{\CurrentBib}

\bibitem [\protect \citeauthoryear {%
Nickel%
\ \BBA {} Kiela%
}{%
Nickel%
\ \BBA {} Kiela%
}{%
{\protect \APACyear {2017}}%
}]{%
nickel}
\APACinsertmetastar {%
nickel}%
\begin{APACrefauthors}%
Nickel, M.%
\BCBT {}\ \BBA {} Kiela, D.%
\end{APACrefauthors}%
\unskip\
\newblock
\APACrefYearMonthDay{2017}{}{}.
\newblock
{\BBOQ}\APACrefatitle {Poincare Embeddings for Learning Hierarchical
  Representations} {Poincare embeddings for learning hierarchical
  representations}.{\BBCQ}
\newblock
\BIn{} \APACrefbtitle {Proceedings of Advances in Neural Information Processing
  Systems.} {Proceedings of advances in neural information processing systems.}
\PrintBackRefs{\CurrentBib}

\bibitem [\protect \citeauthoryear {%
Pa{\c{s}}ca%
\ \BBA {} Van~Durme%
}{%
Pa{\c{s}}ca%
\ \BBA {} Van~Durme%
}{%
{\protect \APACyear {2008}}%
}]{%
pasca-van-durme-2008-weakly}
\APACinsertmetastar {%
pasca-van-durme-2008-weakly}%
\begin{APACrefauthors}%
Pa{\c{s}}ca, M.%
\BCBT {}\ \BBA {} Van~Durme, B.%
\end{APACrefauthors}%
\unskip\
\newblock
\APACrefYearMonthDay{2008}{{\APACmonth{06}}}{}.
\newblock
{\BBOQ}\APACrefatitle {Weakly-Supervised Acquisition of Open-Domain Classes and
  Class Attributes from Web Documents and Query Logs} {Weakly-supervised
  acquisition of open-domain classes and class attributes from web documents
  and query logs}.{\BBCQ}
\newblock
\BIn{} \APACrefbtitle {Proceedings of ACL-08: HLT} {Proceedings of acl-08:
  Hlt}\ (\BPGS\ 19--27).
\newblock
\APACaddressPublisher{Columbus, Ohio}{Association for Computational
  Linguistics}.
\newblock
\begin{APACrefURL} \url{https://aclanthology.org/P08-1003} \end{APACrefURL}
\PrintBackRefs{\CurrentBib}

\bibitem [\protect \citeauthoryear {%
Pennington%
, Socher%
\BCBL {}\ \BBA {} Manning%
}{%
Pennington%
\ \protect \BOthers {.}}{%
{\protect \APACyear {2014}}%
}]{%
glove}
\APACinsertmetastar {%
glove}%
\begin{APACrefauthors}%
Pennington, J.%
, Socher, R.%
\BCBL {}\ \BBA {} Manning, C.%
\end{APACrefauthors}%
\unskip\
\newblock
\APACrefYearMonthDay{2014}{{\APACmonth{10}}}{}.
\newblock
{\BBOQ}\APACrefatitle {{G}lo{V}e: Global Vectors for Word Representation}
  {{G}lo{V}e: Global vectors for word representation}.{\BBCQ}
\newblock
\BIn{} \APACrefbtitle {Proceedings of the 2014 Conference on Empirical Methods
  in Natural Language Processing ({EMNLP})} {Proceedings of the 2014 conference
  on empirical methods in natural language processing ({EMNLP})}\ (\BPGS\
  1532--1543).
\newblock
\APACaddressPublisher{Doha, Qatar}{Association for Computational Linguistics}.
\newblock
\begin{APACrefURL} \url{https://aclanthology.org/D14-1162} \end{APACrefURL}
\newblock
\begin{APACrefDOI} \doi{10.3115/v1/D14-1162} \end{APACrefDOI}
\PrintBackRefs{\CurrentBib}

\bibitem [\protect \citeauthoryear {%
Pinker%
}{%
Pinker%
}{%
{\protect \APACyear {2007}}%
}]{%
pinker}
\APACinsertmetastar {%
pinker}%
\begin{APACrefauthors}%
Pinker, S.%
\end{APACrefauthors}%
\unskip\
\newblock
\APACrefYear{2007}.
\newblock
\APACrefbtitle {The Stuff of Thought} {The stuff of thought}.
\newblock
\APACaddressPublisher{}{Allen Lane}.
\PrintBackRefs{\CurrentBib}

\bibitem [\protect \citeauthoryear {%
Rezende%
, Mohamed%
\BCBL {}\ \BBA {} Wierstra%
}{%
Rezende%
\ \protect \BOthers {.}}{%
{\protect \APACyear {2014}}%
}]{%
rezende14}
\APACinsertmetastar {%
rezende14}%
\begin{APACrefauthors}%
Rezende, D\BPBI J.%
, Mohamed, S.%
\BCBL {}\ \BBA {} Wierstra, D.%
\end{APACrefauthors}%
\unskip\
\newblock
\APACrefYearMonthDay{2014}{}{}.
\newblock
{\BBOQ}\APACrefatitle {Stochastic Backpropagation and Approximate Inference in
  Deep Generative Models} {Stochastic backpropagation and approximate inference
  in deep generative models}.{\BBCQ}
\newblock
\BIn{} \APACrefbtitle {Proceedings of the 31st International Conference on
  Machine Learning} {Proceedings of the 31st international conference on
  machine learning}\ (\BPGS\ 1278--1286).
\PrintBackRefs{\CurrentBib}

\bibitem [\protect \citeauthoryear {%
Rickard%
, Aisbett%
\BCBL {}\ \BBA {} Gibbon%
}{%
Rickard%
\ \protect \BOthers {.}}{%
{\protect \APACyear {2007}}%
}]{%
rickard2007reformulation}
\APACinsertmetastar {%
rickard2007reformulation}%
\begin{APACrefauthors}%
Rickard, J\BPBI T.%
, Aisbett, J.%
\BCBL {}\ \BBA {} Gibbon, G.%
\end{APACrefauthors}%
\unskip\
\newblock
\APACrefYearMonthDay{2007}{}{}.
\newblock
{\BBOQ}\APACrefatitle {Reformulation of the theory of conceptual spaces}
  {Reformulation of the theory of conceptual spaces}.{\BBCQ}
\newblock
\APACjournalVolNumPages{Information Sciences}{177}{21}{4539--4565}.
\PrintBackRefs{\CurrentBib}

\bibitem [\protect \citeauthoryear {%
Rosch%
}{%
Rosch%
}{%
{\protect \APACyear {1973}}%
}]{%
rosch1973natural}
\APACinsertmetastar {%
rosch1973natural}%
\begin{APACrefauthors}%
Rosch, E\BPBI H.%
\end{APACrefauthors}%
\unskip\
\newblock
\APACrefYearMonthDay{1973}{}{}.
\newblock
{\BBOQ}\APACrefatitle {Natural categories} {Natural categories}.{\BBCQ}
\newblock
\APACjournalVolNumPages{Cognitive psychology}{4}{3}{328--350}.
\PrintBackRefs{\CurrentBib}

\bibitem [\protect \citeauthoryear {%
Tull%
}{%
Tull%
}{%
{\protect \APACyear {2021}}%
}]{%
tull2021categorical}
\APACinsertmetastar {%
tull2021categorical}%
\begin{APACrefauthors}%
Tull, S.%
\end{APACrefauthors}%
\unskip\
\newblock
\APACrefYearMonthDay{2021}{}{}.
\newblock
{\BBOQ}\APACrefatitle {A Categorical Semantics of Fuzzy Concepts in Conceptual
  Spaces} {A categorical semantics of fuzzy concepts in conceptual
  spaces}.{\BBCQ}
\newblock
\APACjournalVolNumPages{Proceedings of Applied Category Theory 2021}{}{}{}.
\PrintBackRefs{\CurrentBib}

\bibitem [\protect \citeauthoryear {%
van~den Oord%
, Vinyals%
\BCBL {}\ \BBA {} Kavukcuoglu%
}{%
van~den Oord%
\ \protect \BOthers {.}}{%
{\protect \APACyear {2017}}%
}]{%
discrete-vae}
\APACinsertmetastar {%
discrete-vae}%
\begin{APACrefauthors}%
van~den Oord, A.%
, Vinyals, O.%
\BCBL {}\ \BBA {} Kavukcuoglu, K.%
\end{APACrefauthors}%
\unskip\
\newblock
\APACrefYearMonthDay{2017}{}{}.
\newblock
{\BBOQ}\APACrefatitle {Neural Discrete Representation Learning} {Neural
  discrete representation learning}.{\BBCQ}
\newblock
\BIn{} \APACrefbtitle {Proceedings of Advances in Neural Information Processing
  Systems.} {Proceedings of advances in neural information processing systems.}
\PrintBackRefs{\CurrentBib}

\bibitem [\protect \citeauthoryear {%
Vilnis%
\ \BBA {} McCallum%
}{%
Vilnis%
\ \BBA {} McCallum%
}{%
{\protect \APACyear {2015}}%
}]{%
vilnis}
\APACinsertmetastar {%
vilnis}%
\begin{APACrefauthors}%
Vilnis, L.%
\BCBT {}\ \BBA {} McCallum, A.%
\end{APACrefauthors}%
\unskip\
\newblock
\APACrefYearMonthDay{2015}{}{}.
\newblock
{\BBOQ}\APACrefatitle {Word Representations via Gaussian Embedding} {Word
  representations via gaussian embedding}.{\BBCQ}
\newblock
\BIn{} \APACrefbtitle {Proceedings of {ICLR} 2015.} {Proceedings of {ICLR}
  2015.}
\PrintBackRefs{\CurrentBib}

\bibitem [\protect \citeauthoryear {%
Watters%
, Matthey%
, Borgeaud%
, Kabra%
\BCBL {}\ \BBA {} Lerchner%
}{%
Watters%
\ \protect \BOthers {.}}{%
{\protect \APACyear {2019}}%
}]{%
spriteworld19}
\APACinsertmetastar {%
spriteworld19}%
\begin{APACrefauthors}%
Watters, N.%
, Matthey, L.%
, Borgeaud, S.%
, Kabra, R.%
\BCBL {}\ \BBA {} Lerchner, A.%
\end{APACrefauthors}%
\unskip\
\newblock
\APACrefYearMonthDay{2019}{}{}.
\newblock
\APACrefbtitle {Spriteworld: A Flexible, Configurable Reinforcement Learning
  Environment.} {Spriteworld: A flexible, configurable reinforcement learning
  environment.}
\newblock
\APAChowpublished {https://github.com/deepmind/spriteworld/}.
\newblock
\begin{APACrefURL} \url{https://github.com/deepmind/spriteworld/}
  \end{APACrefURL}
\PrintBackRefs{\CurrentBib}

\end{thebibliography}

\appendix
\renewcommand{\thesection}{\Alph{section}}

\section{The Shapes Dataset}
\label{sec:app_shapes}

The parameters used in the Spriteworld software to generate the Shapes dataset in Section~\ref{sec:shapes}:\\

\noindent
\includegraphics[width=12cm]{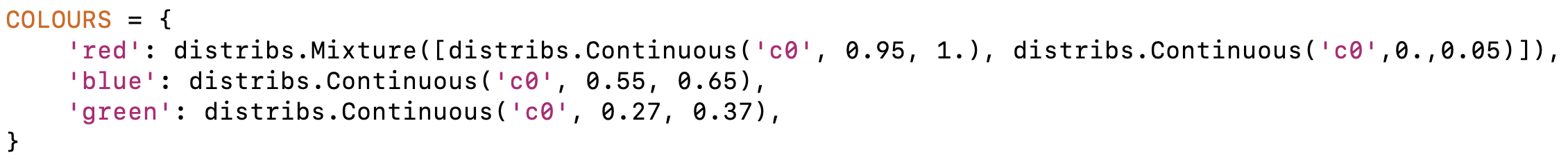}

\noindent
\includegraphics[width=7cm]{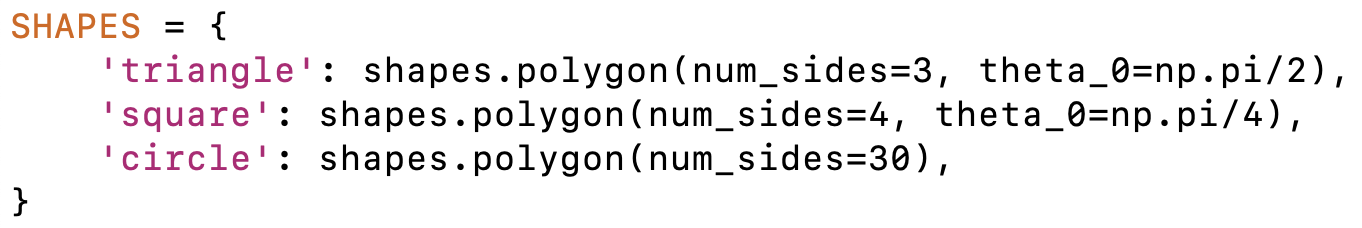}

\noindent
\includegraphics[width=6.5cm]{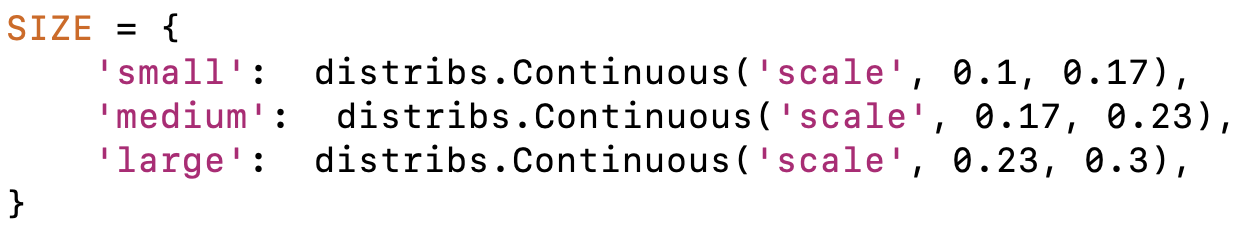}

\noindent
\includegraphics[width=12cm]{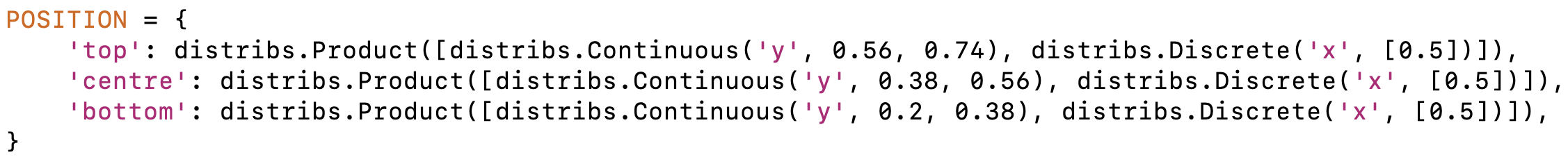}

\noindent
Additional parameters for the \textsc{colour} domain:\\

\noindent
\includegraphics[width=6cm]{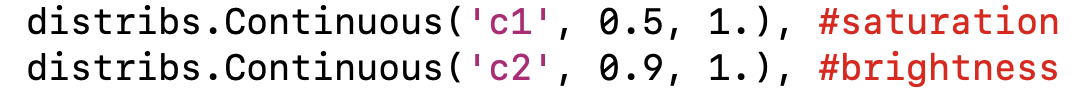}

\subsection{The Extended Colour Dataset}
\label{sec:ext_shapes}

The parameters used in the Spriteworld software to generate the Shapes dataset with more colours in Section~\ref{sec:concept_order}:\\

\noindent
\includegraphics[width=12cm]{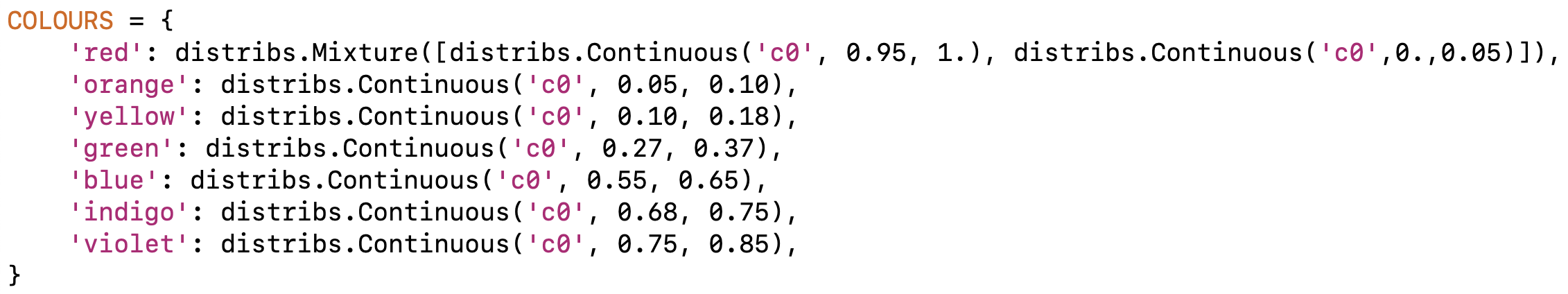}

\section{Neural Architectures and Hyper-parameters}
\label{sec:app_neural_nets}

\begin{center}
\begin{tabular}{@{}lr@{}}
\toprule
image width & 64\\
image height & 64\\
image channels & 3\\
\midrule
CNN kernel size & $4 \times 4$\\
CNN stride & $2 \times 2$\\
CNN layers & 4\\
CNN filters & 64\\
CNN dense layers & 2\\
CNN dense layer size & 256\\
dimensions of latent space & 6\\
\midrule
initialization interval for means of priors & $[-1.0, 1.0]$\\
initialization interval for log-variances of priors & $[-7.0, 0.0]$\\
\midrule
batch size & 32\\
Adam learning rate & $10^{-3}$\\
Adam $\beta_1$ & 0.9 \\
Adam $\beta_2$ & 0.999\\
Adam $\epsilon$ & $10^{-7}$\\
\bottomrule
\end{tabular}
\end{center}


\newpage
\section{Cluster Plots for Dimension 0 for the Rainbow Dataset}
\label{sec:app_rainbow_clusters}
\begin{figure}[b!]
\centering
        \hspace*{-2cm}\includegraphics[scale=0.55]{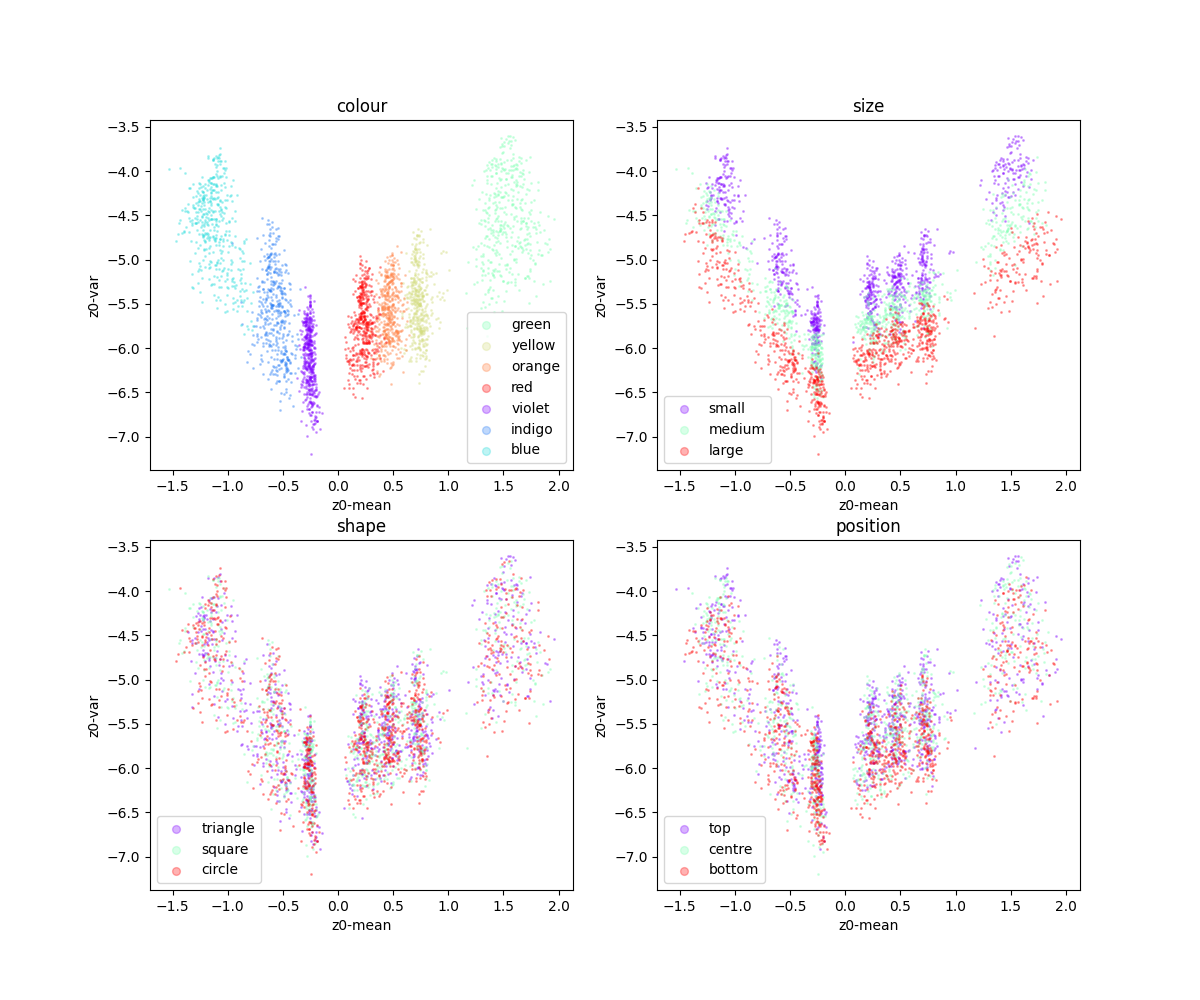}
    \label{fig:clustering_rainbow_colour}
\end{figure}

\newpage
\section{Cluster Plots for 3 \emph{any} Labels}
\label{sec:app_dodgy_clusters}

\begin{figure}[b!]
    \hspace*{-0.5cm}\includegraphics[scale=0.22]{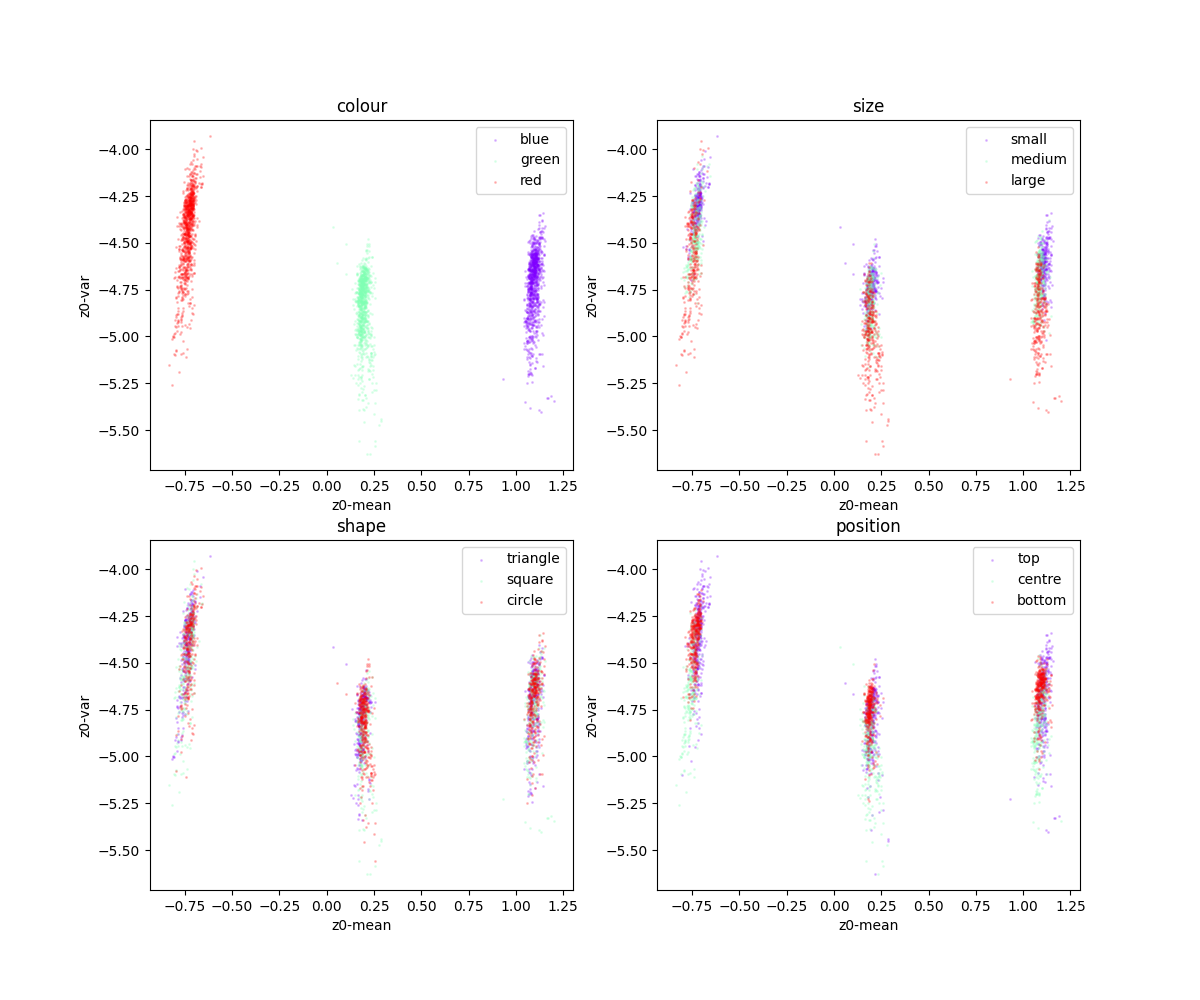}
    \includegraphics[scale=0.22]{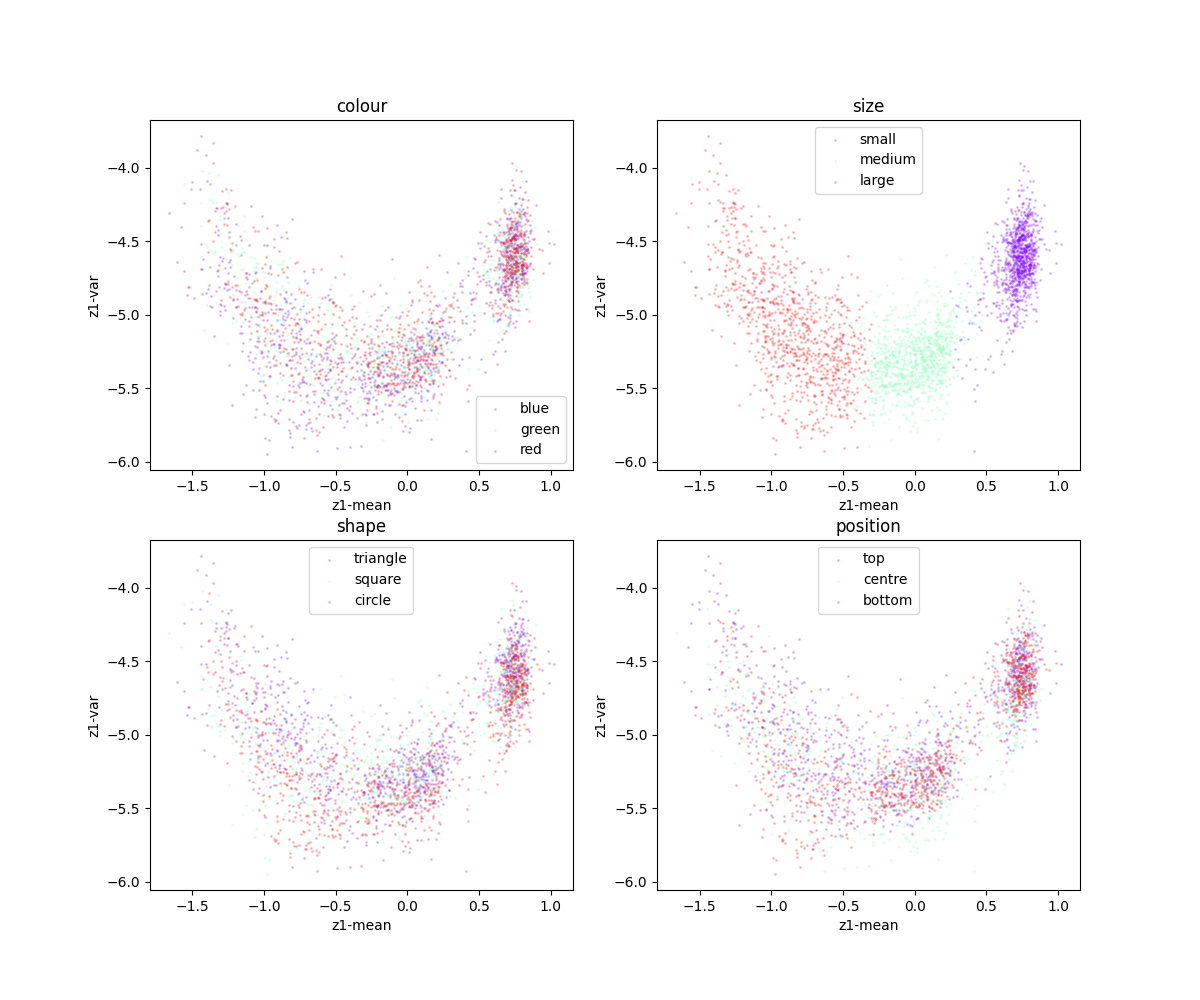}
    \hspace*{-0.5cm}\includegraphics[scale=0.22]{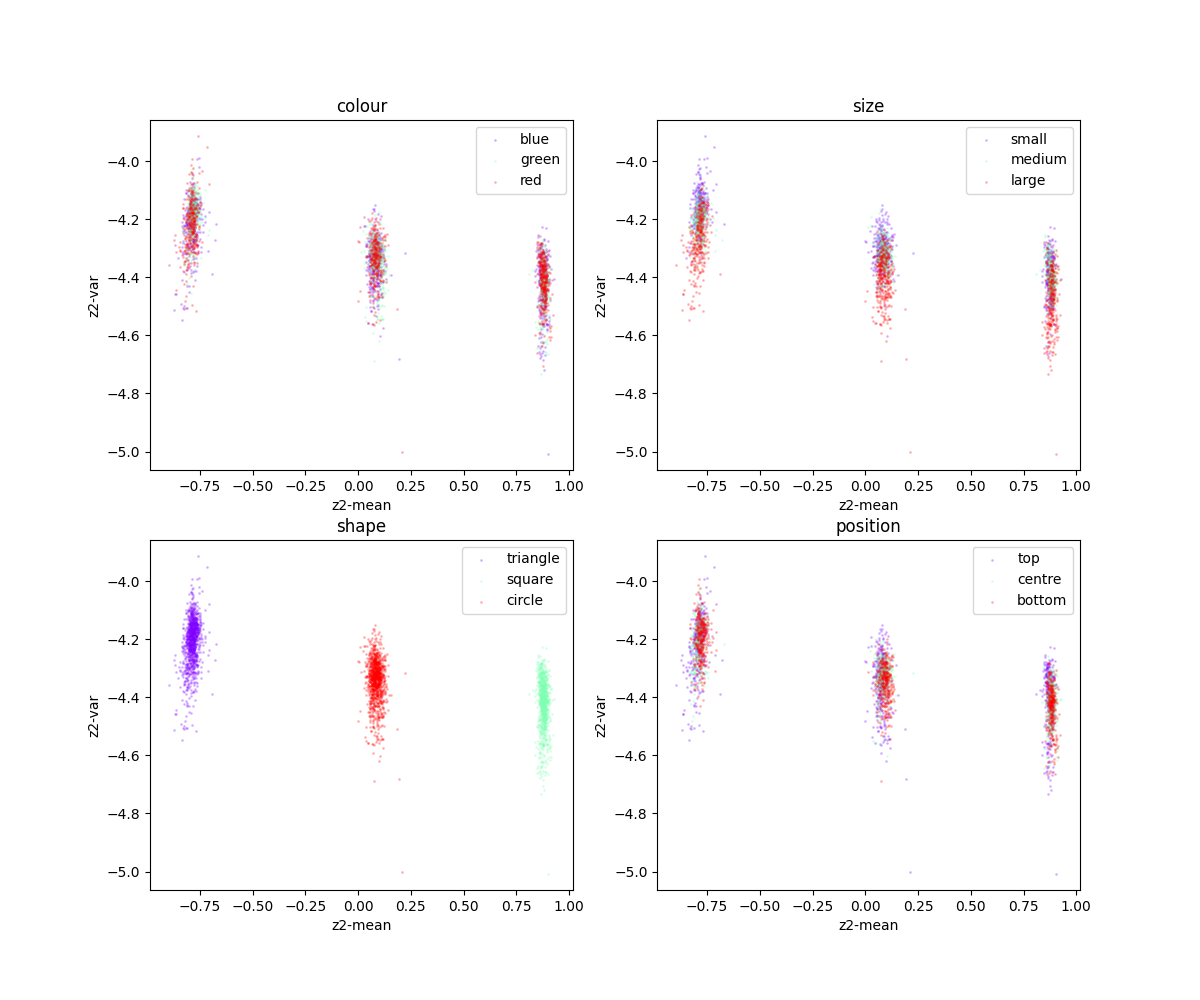}
    \includegraphics[scale=0.22]{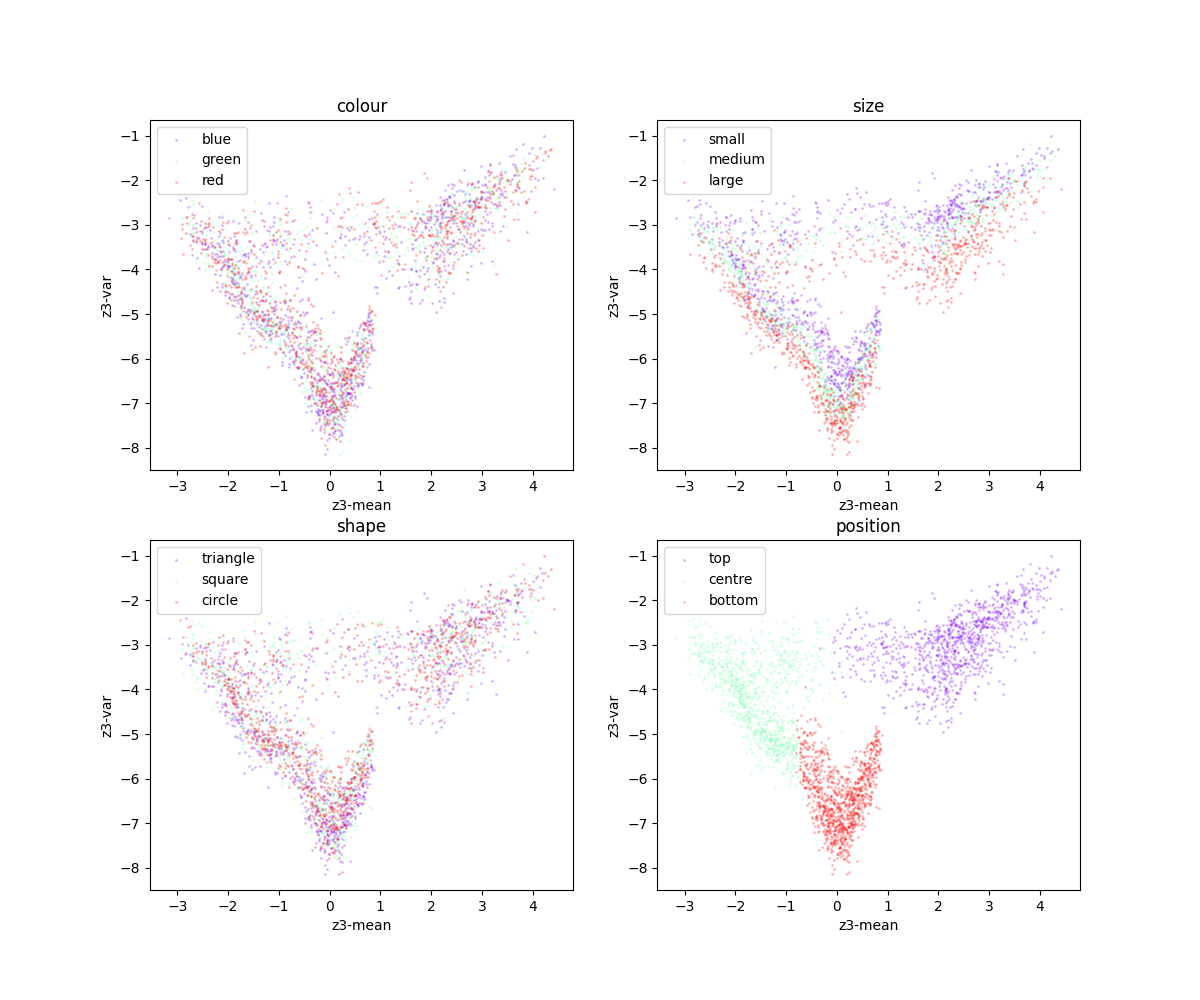}
    \includegraphics[scale=0.22]{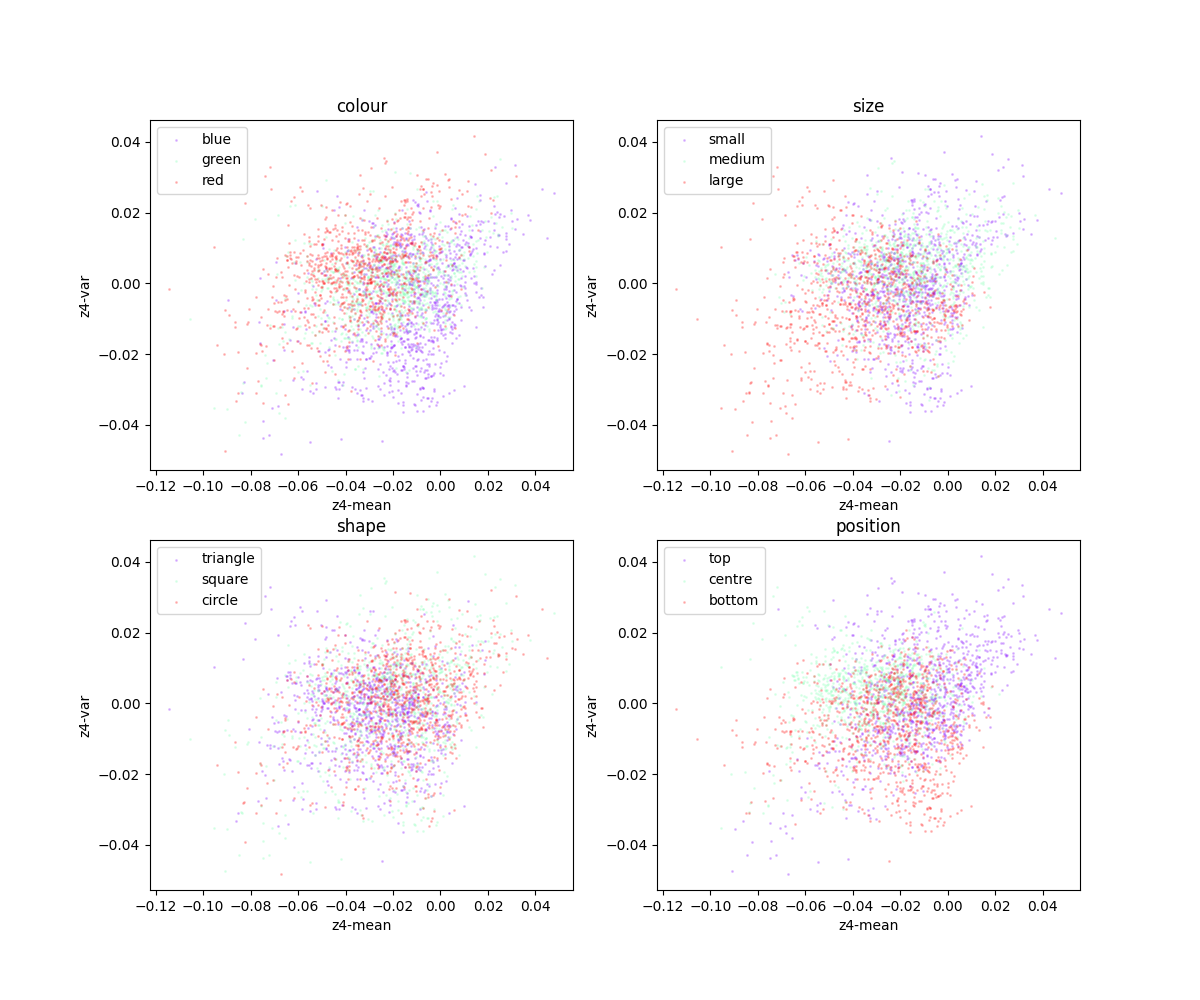}
    \includegraphics[scale=0.22]{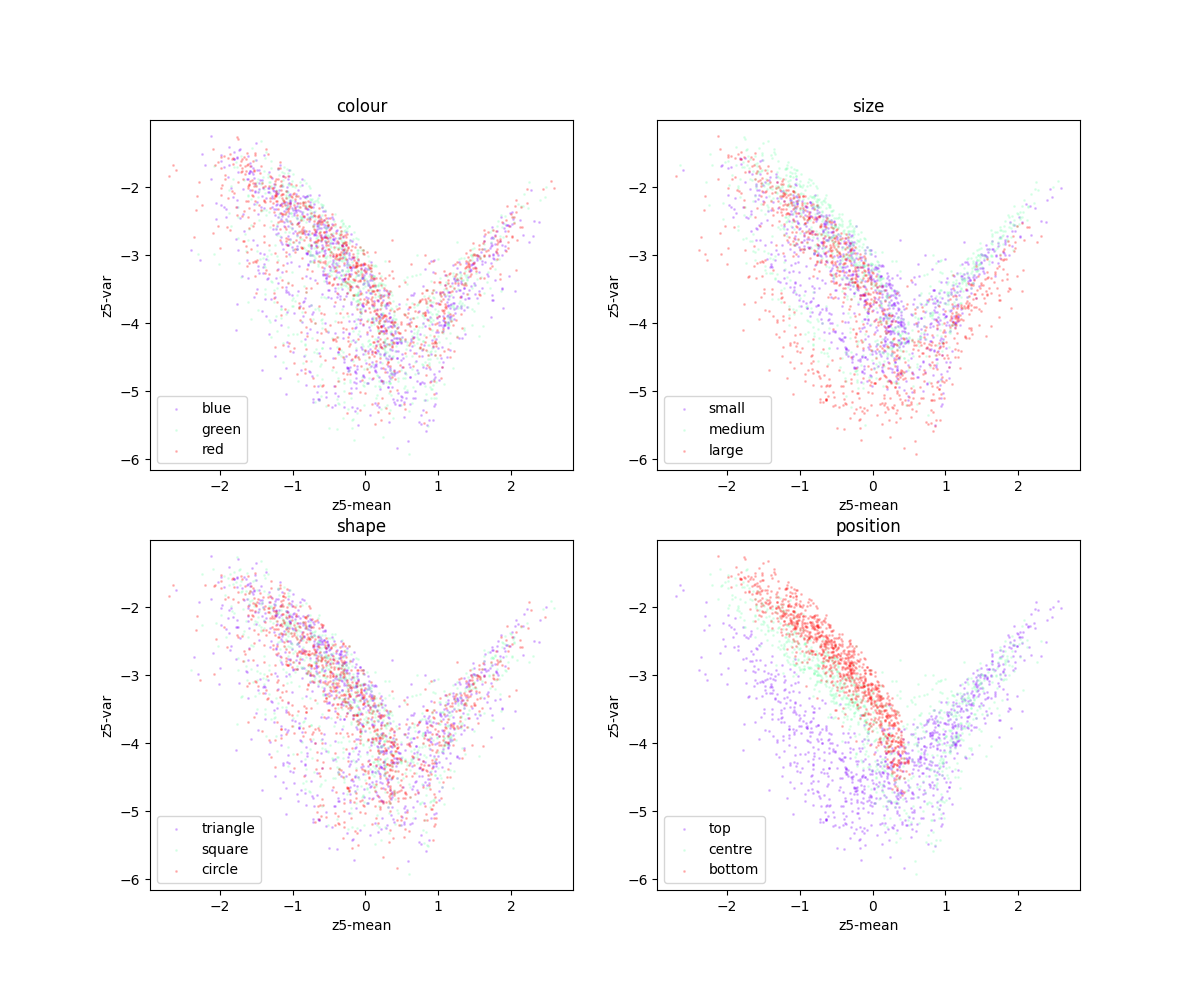}
\end{figure}

\end{document}